\PassOptionsToPackage{table}{xcolor}
\PassOptionsToPackage{pagebackref,breaklinks=true,colorlinks,citecolor=blue,urlcolor=blue,linkcolor=blue,bookmarks=false}{hyperref}
\PassOptionsToPackage{noabbrev,nameinlink,capitalize}{cleveref}
\PassOptionsToPackage{english}{babel}
\AddToHook{package/microtype/after}{\renewcommand{\DisableLigatures}[2][]{}}
\documentclass[onecolumn,numbers]{april_aigc}

\usepackage{latexsym,fix-cm}
\usepackage{array,colortbl,pifont,amsmath,amssymb}
\usepackage{verbatim,float,flafter,fancyhdr,wrapfig}
\definecolor{psyBlueText}{HTML}{3C6787}
\newtcolorbox{psyinsight}{
    colback=aprilbg,colframe=aprilbg,boxrule=0pt,arc=0pt,boxsep=0pt,
    left=6pt,right=6pt,top=5pt,bottom=5pt,before skip=4pt,after skip=5pt
}
\newcommand{\psyInsight}{\textbf{\textcolor{aprilblue}{Key Observation.}}}
\newcommand{\psyChallenge}[1]{\textbf{\textcolor{aprilblue}{#1}}}

\newcommand{\psyChRef}[1]{\textcolor{aprilblue}{(Challenge~#1)}}
\newcommand{\aprcheck}{\ding{51}}
\newcommand{\aprx}{\textcolor{gray!55}{\ding{55}}}
\newcommand{\aprpartial}{\textcolor{gray!70}{\ensuremath{\bullet}}}

\definecolor{linkcolor}{named}{aprilblue}
\definecolor{urlcolor}{RGB}{255,105,180}
\definecolor{citecolor}{RGB}{66,168,235}
\hypersetup{colorlinks=true,linkcolor=linkcolor,urlcolor=urlcolor,citecolor=citecolor}
\crefname{section}{Sec.}{Secs.}
\Crefname{section}{Section}{Sections}
\crefname{table}{Tab.}{Tabs.}
\Crefname{table}{Table}{Tables}
\crefname{equation}{Eq.}{Eqs.}
\Crefname{equation}{Equation}{Equations}
\crefname{figure}{Fig.}{Figs.}
\Crefname{figure}{Figure}{Figures}
\titleformat*{\section}{\color{aprilblue}\Large\sffamily\bfseries}
\titleformat*{\subsection}{\color{aprilblue}\large\sffamily\bfseries}
\titleformat*{\subsubsection}{\color{aprilblue}\normalsize\sffamily\bfseries}
\DeclareCaptionFormat{custom}{{\color{aprilblue}\sffamily\textbf{#1 #2}} #3}
\renewcommand{\title}[1]{\def\titlelist{{\fontsize{20pt}{28pt}\selectfont\sffamily\bfseries #1}}}

\title{PsycheMME: A Psychology-Informed Benchmark
for Social Understanding and Interaction in Videos}

\author[1,2]{Heran Hu}
\author[1]{Yuhang Dong}
\author[1]{Zhucun Xue}
\author[1]{Kaiwei Chen}
\author[1]{Yong Liu}
\author[1\raisebox{-0.2em}{\includegraphics[height=0.85em]{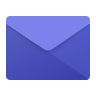}}]{Jiangning Zhang}
\affiliation[1]{Zhejiang University, APRIL Lab}
\affiliation[2]{China University of Geosciences}
\coverdate{\today}
\covercorrespondence{\email{186368@zju.edu.cn}}
\coversourcecode{https://github.com/psychemme/PsycheMME}
\coverdataset{https://huggingface.co/datasets/psychemme/PsycheMME}
\coverproject{https://psychemme.github.io/}
\abstract{Understanding human needs and selecting responses require distinct judgments that aggregate accuracy can obscure. We introduce \textbf{PsycheMME}, a video benchmark informed by psychology, with eight capabilities spanning grounding, states, needs, interaction decisions, memory, and risk. Drawing on Social Information Processing and the Perception-Action Cycle, this organization separates human understanding from response selection. We construct 1,000 questions from 630 episodes through independent answering with proposed references withheld and expert review. Our \textbf{Reasoning-Verified} protocol weights answer credit by the support and logical consistency of explicit rationales against annotated cues. Across 17 models, Timing has the lowest Verified scores and credit retention, with weaker support judgments for correct Timing answers. Models with similar Need scores still differ in answer completeness on open questions. Input controls and SFT/GRPO examine context dependence and adaptation. Evaluating answers alongside rationales locates where models fail across understanding, response selection, and justification.}

\begin{document}
\maketitle
\thispagestyle{plain}

\begin{figure*}[t!]
    \centering
    \includegraphics[width=\textwidth]{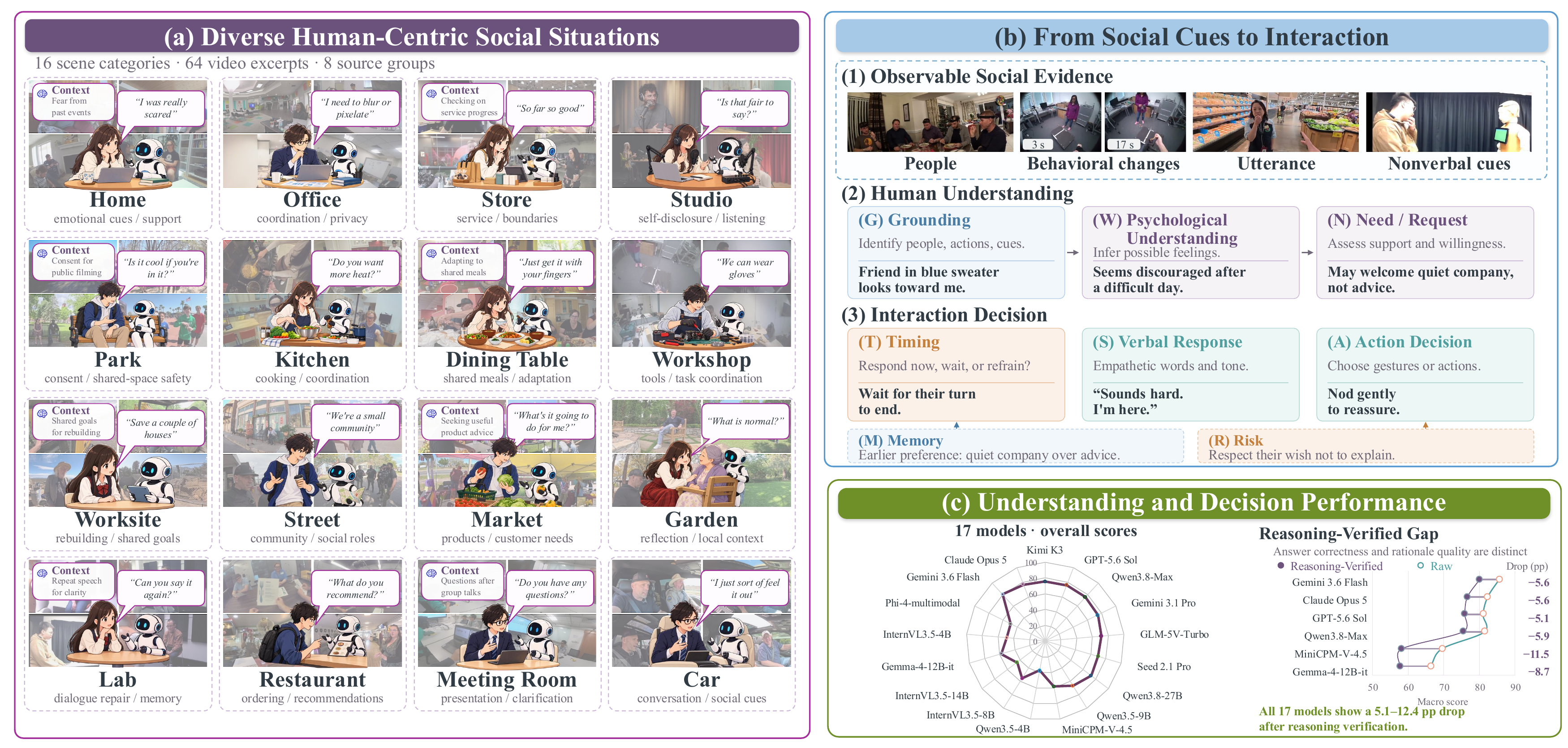}
    \vspace{-20pt}
    \caption{
        \textbf{Overview of PsycheMME.}
        (a) Social situations involving people across
        16 scene categories.
        (b) Eight diagnostic capabilities covering
        social cues, psychological understanding,
        interaction decisions, memory, and risk.
        (c) Overall results for 17 models, showing
        performance differences and
        gaps after rationale verification.
    }
    \label{fig:method}
\end{figure*}

\section{Introduction}
\label{sec:introduction}

Multimodal Foundation Models have expanded from image and text
understanding to video and audio
\cite{alayrac2022flamingo,xu2025qwen25omni}.
Their evaluation spans video understanding and increasingly
examines people in egocentric and social settings
\cite{grauman2022ego4d,mangalam2023egoschema,dang2025ecbench,liu2025humanmme,cai2025humanvideomme}.
Here, recognizing actions is not enough: models must
interpret social cues, infer psychological states and needs,
and make interaction judgments appropriate to the context
\cite{mathur2025socialgenome,qin2026humansense}.
Inferring latent states from observable cues is a central problem in
social cognition~\cite{premack1978tom}. We ask:
\emph{Can multimodal foundation models understand human
states and needs from observable cues and make
appropriate interaction judgments?}

Existing benchmarks and embodied systems address aspects
of this question, from perception and social reasoning
\cite{mathur2025socialgenome,kong2026siv}
to personalization and embodied decision making
\cite{yang2025egolife,driess2023palme,kim2024openvla,zhang2026mindpower}.
Broader benchmarks~\cite{qin2026humansense,ong2026human} still differ
in capability definitions and assessment criteria, making failures
difficult to compare.
Aggregate accuracy leaves two diagnostic questions unanswered:
where do models struggle between social observation and response selection,
and when do they select the right option without adequate cue support?

\begin{psyinsight}
\psyInsight\
A correct answer does not guarantee a well supported explanation.
Figure~\ref{fig:failure_case} shows a correct response strategy justified
with cues from the wrong time window, alongside errors in timing,
preference use, and need interpretation
(Sections~\ref{sec:capability_taxonomy}
and~\ref{sec:evaluation_protocol}).
\end{psyinsight}

These requirements raise three challenges:

\noindent
\psyChallenge{Challenge 1: Understanding and response selection}
(Section~\ref{sec:capability_taxonomy}).
Evaluation must assess state inference alongside timing, language,
and action choices, with memory and boundaries as context.

\par\noindent
\psyChallenge{Challenge 2: Psychological constructs and observable cues}
(Section~\ref{sec:benchmark_construction}).
Mental states and support needs are latent. Questions must anchor
interpretations in observable visual and conversational cues
without unsupported attribution.

\par\noindent
\psyChallenge{Challenge 3: Cue support beyond option matching}
(Section~\ref{sec:evaluation_protocol}).
Assessment must check cue support and logical consistency of explicit
rationales, even when the selected option is correct.

To address these challenges, we introduce
\textbf{PsycheMME}, a benchmark informed by psychology
for multimodal evaluation of human understanding and interaction.
Drawing on Social Information Processing
\cite{crick1994sip} and the Perception-Action Cycle
\cite{fuster2004perception}, we define eight
diagnostic dimensions spanning perceptual grounding,
psychological understanding, needs, interaction decisions,
memory, and risk~\psyChRef{1}.
Independent answering with proposed references withheld
and expert review ground psychological QA in observed
cues~\psyChRef{2}.
Reasoning-Verified assesses answer correctness, rationale
coverage, Evidence Support, and Logical Integrity,
separating response success from justification quality~\psyChRef{3}.

Our main contributions are as follows:

\begin{itemize}
\item \textbf{A diagnostic benchmark informed by psychology.}
PsycheMME contains 1,000 questions from 630 episodes across eight
capabilities: understanding (G, W, N), response selection (T, S, A),
and contextual regulation (M, R).

\item \textbf{Independent curation and rationale scoring.}
Agent roles isolate independent answering from reference proposals,
followed by unanimous expert review. \textbf{Reasoning-Verified}
assesses explicit rationales for cue support and logical coherence
alongside answer correctness.

\item \textbf{Diagnostic findings across 17 MLLMs.}
Weak or contradicted support labels occur in 49.74\% of correct Timing
answers, while open Need full credit rates differ by 40 percentage points
between Gemini 3.6 Flash and Claude Opus 5. Voting, input, history,
and SFT/GRPO analyses examine answer selection and adaptation.
\end{itemize}

\begin{table*}[tp]
    \centering
    \caption{
        \textbf{Comparison with Existing Multimodal Benchmarks of Human Understanding and Interaction.}
    }
    \label{tab:benchmark_comparison}

    \renewcommand{\arraystretch}{1.04}
    \setlength{\tabcolsep}{7.0pt}
    \resizebox{0.98\textwidth}{!}{%
        \begin{tabular}{lccccccccccc}
        \toprule
        Benchmark & Videos/Clips & Prompt & Ego
        & G & W & N & T & S & A & M & R \\
        \midrule

        HumanVideo-MME~\cite{cai2025humanvideomme}
        & 1,200 & 8,700 & \aprx
        & \aprcheck & \aprcheck & \aprx & \aprx
        & \aprx & \aprx & \aprx & \aprx \\

        Social Genome~\cite{mathur2025socialgenome}
        & 272 & 1,486 & \aprx
        & \aprcheck & \aprcheck & \aprx & \aprx
        & \aprx & \aprx & \aprx & \aprx \\

        Human Behavior Atlas~\cite{ong2026human}
        & 35,046 & 101,964 & \aprx
        & \aprpartial & \aprcheck & \aprx & \aprx
        & \aprx & \aprx & \aprx & \aprx \\

        \midrule

        HumanSense~\cite{qin2026humansense}
        & - & 3,882 & \aprx
        & \aprcheck & \aprcheck & \aprpartial & \aprx
        & \aprcheck & \aprcheck & \aprpartial & \aprx \\

        SIV-Bench~\cite{kong2026siv}
        & 2,792 & 5,455 & \aprx
        & \aprcheck & \aprcheck & \aprx & \aprx
        & \aprx & \aprx & \aprx & \aprx \\

        R$^3$-Bench~\cite{niu2026readtheroom}
        & 312 & 4,840+316 & \aprx
        & \aprcheck & \aprcheck & \aprpartial & \aprx
        & \aprx & \aprx & \aprx & \aprx \\

        ToM-Video~\cite{tomvideo2026}
        & 4,666 & 41,206 & \aprx
        & \aprcheck & \aprcheck & \aprpartial & \aprx
        & \aprx & \aprx & \aprx & \aprx \\

        MindPower~\cite{zhang2026mindpower}
        & 590 & 590 & \aprx
        & \aprcheck & \aprcheck & \aprcheck & \aprx
        & \aprx & \aprcheck & \aprx & \aprx \\

        SHREC~\cite{lee2026shrec}
        & $\sim$400 & - & \aprx
        & \aprcheck & \aprcheck & \aprx & \aprcheck
        & \aprcheck & \aprpartial & \aprcheck & \aprcheck \\

        EgoPro-Bench~\cite{ran2026egopro}
        & 2,400 & - & \aprcheck
        & \aprx & \aprcheck & \aprcheck & \aprcheck
        & \aprcheck & \aprx & \aprcheck & \aprx \\

        RobotEQ-Video~\cite{che2026roboteq}
        & 2,036 & - & \aprcheck
        & \aprx & \aprpartial & \aprpartial & \aprx
        & \aprx & \aprcheck & \aprx & \aprcheck \\

        SocialReasonBench~\cite{huang2026socialreasonbench}
        & 532 & 532 & \aprx
        & \aprx & \aprcheck & \aprx & \aprx
        & \aprx & \aprx & \aprx & \aprx \\

        \midrule
        \rowcolor{aprilblue!20}
        \textbf{PsycheMME}
        & 630 & 1,000 & \aprcheck
        & \aprcheck & \aprcheck & \aprcheck & \aprcheck
        & \aprcheck & \aprcheck & \aprcheck & \aprcheck \\
        \bottomrule
        \end{tabular}%
    }

    \vspace{2pt}
    \begin{minipage}{0.98\textwidth}
        \footnotesize
        \textit{Notes.}
        G: Grounding; W: Psychological Understanding;
        N: Need; T: Interaction Timing;
        S: Verbal Response; A: Embodied Action;
        M: Memory; R: Risk.
        \aprcheck\ explicit;
        \aprpartial\ restricted;
        \aprx\ unestablished evaluation.
        Counts use each source's reported video/clip/episode and
        question/instance units. ToM-Video reports 4,666 clips from
        1,487 source videos.
        Human Behavior Atlas counts include all training, validation,
        and test samples across text, audio, and video.
        Criteria: Appendix~\ref{app:benchmark_comparison}.
    \end{minipage}
\end{table*}

\section{Related Work}
\label{sec:related_work}

\subsection{Multimodal Foundation Models}
\label{sec:rw_mfm}

Multimodal Foundation Models extend learning from images and
text~\cite{alayrac2022flamingo} to video, audio,
and reasoning over long contexts
\cite{wang2025internvl35,xu2025qwen25omni}.
Magma further connects multimodal understanding to action
\cite{yang2025magma}.
Beyond scene perception, PersonaVLM explores personalized memory,
MindPower connects reasoning about human states to embodied actions,
and FAMER addresses implicit user desires through communication and memory
\cite{nie2026personavlm,zhang2026mindpower,
wang-etal-2026-communication}.
These systems motivate evaluation of how social observations,
personal history, and inferred needs inform response choices.

\subsection{Multimodal Benchmarks of Human Understanding and Interaction}
\label{sec:rw_social_response}

Multimodal benchmarks assess understanding and interaction.
PCA-Bench evaluates how perception and cognition inform action,
localizing errors within this chain
\cite{chen2024pcabench};
Human Behavior Atlas integrates psychological and social tasks
\cite{ong2026human};
HumanSense covers perception, contextual understanding, and empathetic
responses~\cite{qin2026humansense}.
Social Genome evaluates reasoning traces grounded in
multimodal social cues~\cite{mathur2025socialgenome}.
Other benchmarks assess video understanding
\cite{mangalam2023egoschema,cheng2024egothink,plizzari2025egotempo},
embodied decision making~\cite{yang2025embodiedbench},
egocentric memory and assistance
\cite{dang2025ecbench,yang2025egolife,xiao2026myego},
human perception~\cite{liu2025humanmme,cai2025humanvideomme},
and social interaction and reasoning
\cite{tomvideo2026,huang2026socialreasonbench,jin2024mmtom,
niu2026readtheroom,kong2026siv,lee2026shrec}.
EgoPro-Bench and LiveProBench examine interaction timing
\cite{ran2026egopro,du2026proactivebench},
while related efforts assess social appropriateness~\cite{che2026roboteq},
implicit norms~\cite{zhao2026normact},
and embodied safety~\cite{ying2026agentsafe}.
PsycheMME brings these judgments into a shared setting for video QA,
distinguishing response timing, content, and action while assessing
the support for explicit explanations.

\subsection{Evaluation Informed by Psychology}
\label{sec:rw_psychology}

Models of affect, Theory of Mind (ToM), and social cognition in direct
interaction inform the assessment of human states
\cite{russell1980circumplex,premack1978tom,schilbach2013toward}.
Social Information Processing (SIP) links cue encoding and interpretation
to goal clarification and response selection~\cite{crick1994sip}.
Research on empathy, interpersonal emotion regulation, and matching
support to circumstances informs response assessment
\cite{davis1983empathy,zaki2013interpersonal,cutrona1992support}.
FANToM, EmoBench, and ToMBench adapt psychological constructs
for model evaluation~\cite{kim2023fantom,sabour2024emobench,chen2024tombench}.
These perspectives inform PsycheMME's task definitions;
its rubric for cue support and logic links answer credit to
contextually supported explanations.
LLM judges and evaluation with rubrics assess open responses using
explicit criteria~\cite{zheng2023judge,liu2023geval}.

\section{Methodology}
\label{sec:methodology}

\subsection{Task Formulation}
\label{sec:task_formulation}

\begin{figure*}[!htbp]
    \centering
    \includegraphics[width=\textwidth]{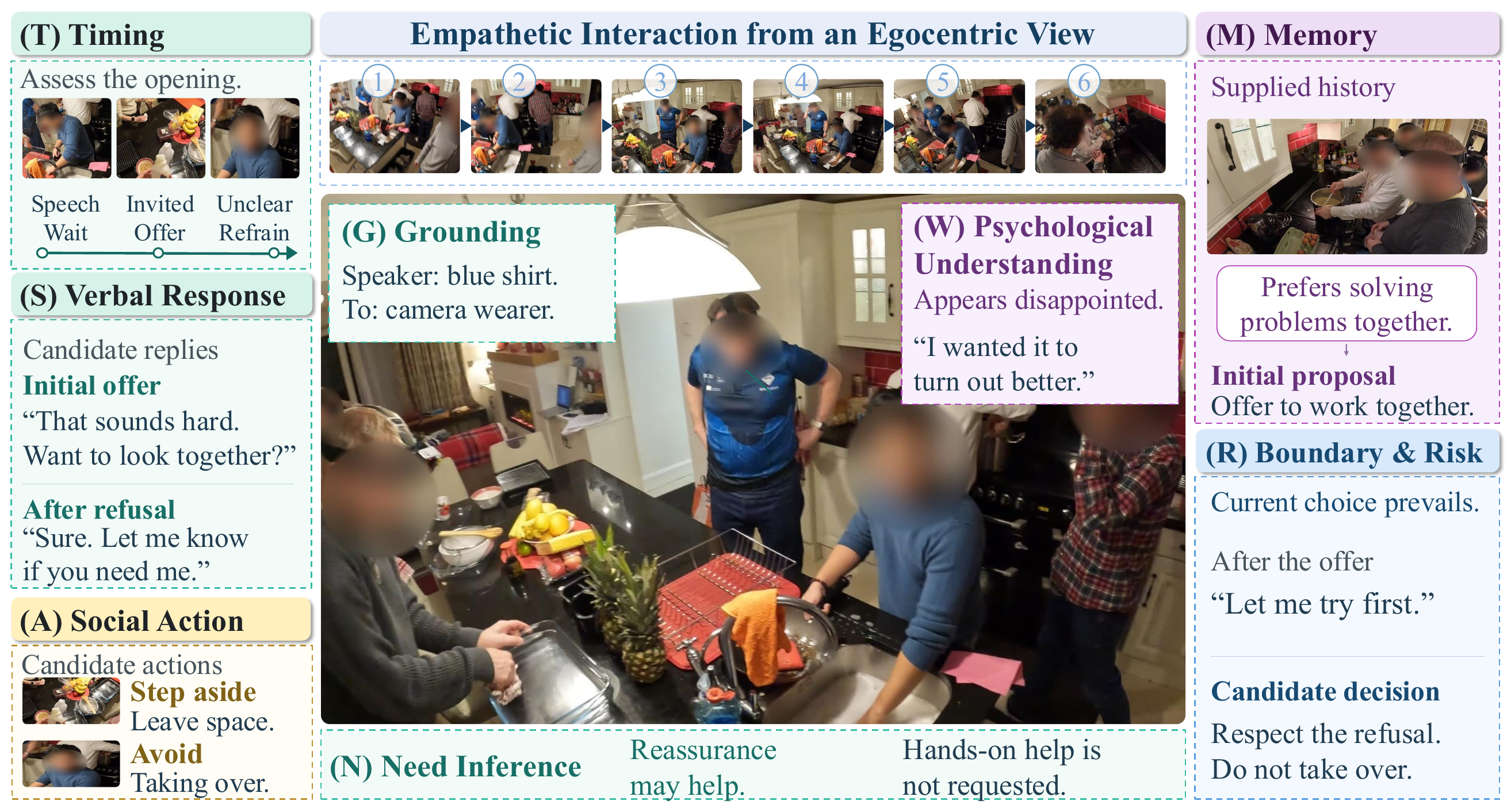}
    \vspace{-20pt}
    \caption{
    \textbf{Illustrative scenarios and capabilities in PsycheMME.}
    The benchmark spans human understanding, interaction
    decisions, memory, and social constraints.
    The illustrated scene is adapted from
    CASTLE~\cite{rossetto2025castle}.
    }
    \label{fig:teaser}
\end{figure*}

PsycheMME evaluates how models perceive social cues, infer mental states,
and select responses in recorded interactions.
Each question targets an observation, interpretation,
or response choice grounded in the supplied context.
Given multimodal observations $X$, question $q$, and optional
interaction history $h$, the model predicts
\begin{equation}
    (\hat{y},\,\hat{\boldsymbol{\rho}}) = f_\theta(X,q,h),
    \label{eq:task_formulation}
\end{equation}
where $\hat{y}$ is the answer and $\hat{\boldsymbol{\rho}}$ contains
structured rationales associated with eight capabilities:
Grounding (G), Psychological Understanding (W), Need (N),
Timing (T), Verbal Response (S), Action (A), Memory (M), and Risk (R)
(Section~\ref{sec:capability_taxonomy}).

\subsection{Capability Taxonomy Informed by Psychology}
\label{sec:capability_taxonomy}

Social Information Processing~\cite{crick1994sip} and the
Perception-Action Cycle~\cite{fuster2004perception} motivate
three diagnostic domains (Fig.~\ref{fig:teaser}):
human understanding (G, W, N), interaction decisions
(T, S, A), and context and constraints (M, R).
Detailed definitions and requirements for observed cues appear in
Appendix~\ref{app:taxonomy_details}.

\paragraph{Perceptually Grounded Mentalizing.}
\textbf{Social Grounding (G)} associates speech, gaze, gestures,
and actions with participants and interaction relations.
\textbf{Psychological Understanding (W)} uses multimodal cues and context
to infer psychological states, possible causes, and changes.
\textbf{Need Inference (N)} identifies desired support,
distinguishing explicit requests, inferred needs, and willingness
to receive help~\cite{cutrona1992support,bolger2000invisible}.

\paragraph{Situated Empathic Responding.}
\textbf{Interaction Timing and Attunement (T)} judges whether and when to
respond, wait, listen, or refrain~\cite{sacks1974turntaking}.
\textbf{Empathic Verbal Responding (S)} selects response content,
strategy, and tone~\cite{cutrona1992support,davis1983empathy,zaki2013interpersonal}.
\textbf{Embodied Social Action (A)} selects physical or nonverbal
behavior appropriate to the person's state, need, and context.

\paragraph{Contextual and Normative Modulation.}
\textbf{Interaction and Profile Memory (M)} applies prior exchanges
and individual preferences to current judgments~\cite{nie2026personavlm}.
\textbf{Risk, Boundary, and Autonomy Regulation (R)} assesses potential
harm, drawing on embodied safety evaluation~\cite{ying2026agentsafe},
and additionally considers consent, boundaries, and autonomy
in proposed interventions~\cite{brown1987politeness,ryan2000self}.

\subsection{Scalable Benchmark Construction}
\label{sec:benchmark_construction}

\begin{figure*}[!htbp]
    \centering
    \includegraphics[width=\textwidth]{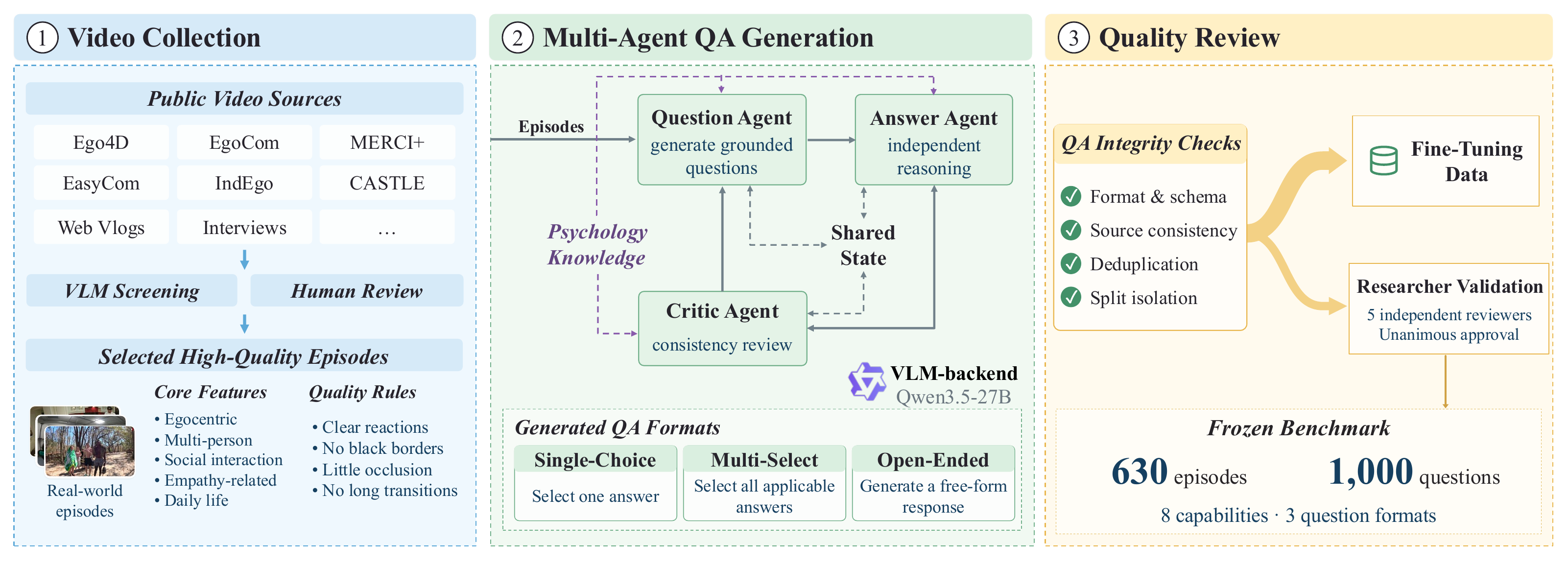}
    \vspace{-20pt}
    \caption{
    \textbf{Question construction framework of PsycheMME.}
    Video curation, independent agent answers with proposed answers
    withheld, verification, iterative refinement, and human quality control.
    }
    \label{fig:construction}
\end{figure*}

\paragraph{Video Collection.}
We collect videos of social interactions from six public datasets
(Ego4D~\cite{grauman2022ego4d}, EgoCom~\cite{northcutt2020egocom},
MERCI+ (the extended MERCI release~\cite{althubyani2025merci}), CASTLE~\cite{rossetto2025castle},
EasyCom~\cite{donley2021easycom}, and IndEgo~\cite{chavan2025indego}) and public
vlogs, interviews, and recordings of daily life. Human screening
assisted by VLMs retains clips with observable social cues and
adequate visual quality.

\paragraph{Multi-Agent QA Generation.}
The question construction framework in Fig.~\ref{fig:construction} combines
question generation, independent answering, verification, and refinement.
Separating question proposal from independent answering reduces the
risk of accepting a proposal based on its own justification.
The \textbf{Question Agent} proposes a question, candidate
answer, options where applicable, and supporting cues.
The \textbf{Answer Agent} examines the video again using only
the question, options, and public context, without access to the
proposed answer, cues, or private rationale. It independently answers
or flags ambiguity and insufficient support.
The \textbf{Critic Agent} checks both paths for answerability, cue
support, and logical coherence. Proposal defects return to the
Question Agent; reasoning defects trigger another examination.
\textbf{Psychology Knowledge} guides the agents' construct alignment and
inference boundaries; items failing after two refinement rounds are discarded.
A \textbf{Shared State} with access control based on agent roles enforces
information isolation and tracks verification feedback, refinement
history, and taxonomy coverage to guide generation toward
underrepresented capabilities (Appendix~\ref{app:construction_details}).

\paragraph{Quality Review.}
For benchmark construction, five researchers
with psychology backgrounds independently review
question validity, answer correctness, and
psychological consistency. Only unanimously
approved items enter the frozen benchmark.
Data for fine tuning are separately filtered and
kept disjoint from benchmark items.

\FloatBarrier
\subsection{Benchmark Composition and Statistics}
\label{sec:benchmark_statistics}

\begin{figure*}[!htbp]
    \centering
    \includegraphics[width=\textwidth]{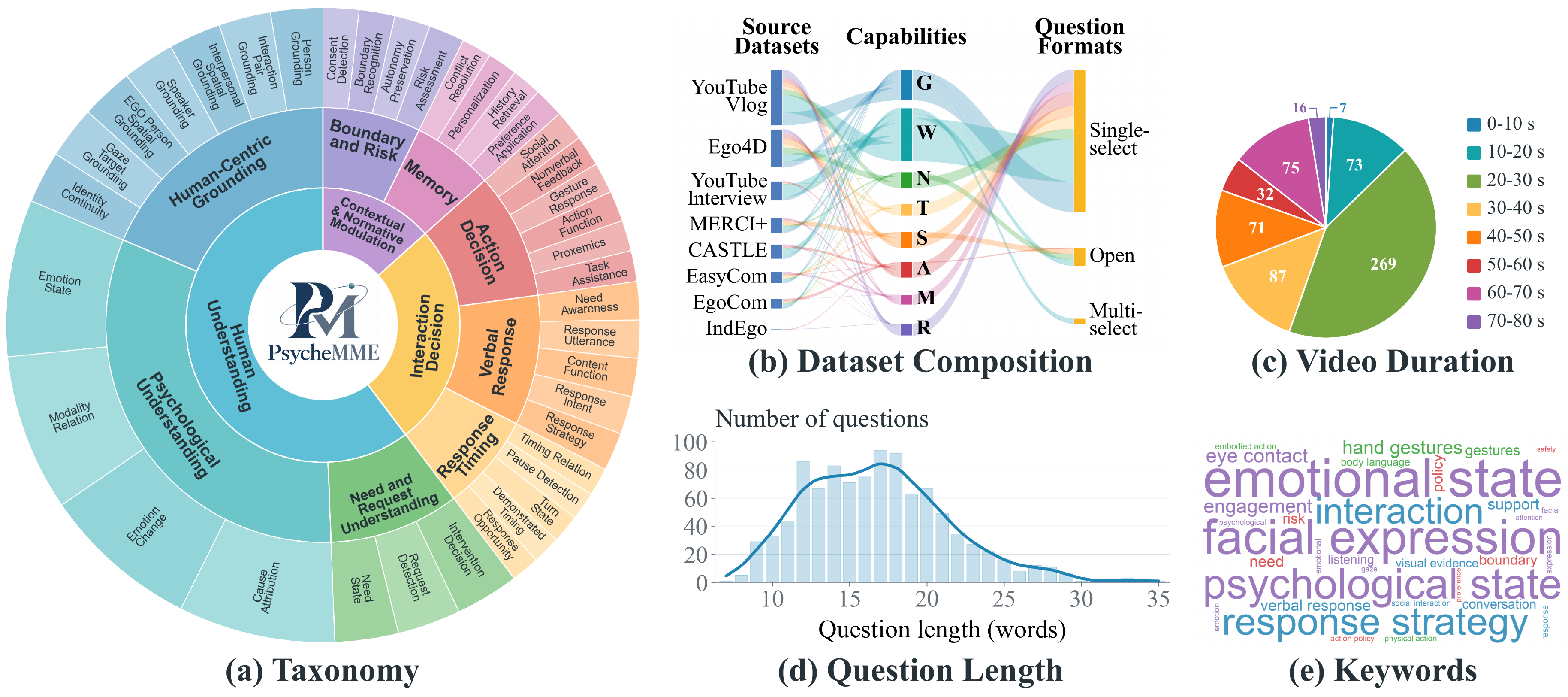}
    \vspace{-20pt}
    \caption{
    \textbf{PsycheMME statistics.}
    (a) Eight capabilities and their subcategories.
    (b) Question distribution across source groups,
    capabilities, and formats.
    (c) Distribution of video durations.
    (d) Distribution of question lengths in words.
    (e) Frequent keywords in benchmark questions.
    }
    \label{fig:benchmark_statistics}
\end{figure*}

PsycheMME contains 1,000 questions from 630 episodes across eight
capabilities: 890 choice questions and 110 open questions.
Choice questions include selection of one or several options;
option letters identify candidates within each question.
Uneven capability counts motivate macro averaging; distributions appear
in Fig.~\ref{fig:benchmark_statistics} and
Appendix~\ref{app:composition_details}.

\subsection{Evaluation Protocol}
\label{sec:evaluation_protocol}
We report \textbf{Raw Answer} and
\textbf{Reasoning-Verified} scores.
Choice answers match the reference option or complete option set exactly;
a fixed semantic judge evaluates open answers.
Reasoning-Verified additionally considers rationale
coverage, support from reference cues, and
logical integrity:
\begin{equation}
    s_i = a_i c_i e_i \ell_i / \lambda^2,
    \label{eq:verified_score}
\end{equation}
where $a_i\in[0,1]$ denotes answer correctness,
$c_i\in\{0,1\}$ indicates an informative rationale, and
$e_i,\ell_i\in[0,\lambda]$ measure support from
reference cues and logical integrity, respectively.
With $\lambda=1.2$, full Verified credit requires a correct answer,
an informative rationale, strong cue support, and fully coherent logic
($e_i=\ell_i=1.2$). Normal support and acceptable logic
($e_i=\ell_i=1.0$) receive $1/1.44\approx69.44\%$ credit.
The rubric grades justification quality beyond answer correctness.
Both metrics are macro averages across eight capabilities
(Appendices~\ref{app:judge_rubric} and~\ref{app:scoring_details}).

\subsection{SFT and GRPO on PsycheMME}
\label{sec:task_specific_post_training}

We apply SFT followed by GRPO to Qwen3.5-4B using
LoRA~\cite{hu2022lora} on the language backbone, with the visual encoder and
multimodal aligner frozen. SFT trains for one epoch
on 3,657 video QA examples with 32 sampled frames,
ASR, and available interaction history. Targets
contain the answer and one annotated observation;
other rationale fields are null. Training uses AdamW~\cite{loshchilov2019decoupled}.

GRPO~\cite{shao2024deepseekmath} starts from SFT and samples eight responses
per question from 2,842 single-choice examples.
Its binary reward requires a correct answer that satisfies the
specified response requirements:
\begin{equation}
    r(o) = \mathbf{1}[\mathrm{correct}(o)]
    \cdot \mathbf{1}[\mathrm{valid}(o)],
    \label{eq:grpo_reward}
\end{equation}
We use clipped GRPO with advantages normalized within each group
and KL regularization against the fixed SFT reference,
excluding groups with zero variance following the filtering criterion
used in DAPO~\cite{yu2025dapo}. Training performs
32 updates over 256 distinct questions.
Reasoning verification is used only for evaluation;
details appear in Appendix~\ref{app:post_training}.

\section{Experiments}
\label{sec:experiments}

\subsection{Experimental Setup}
\label{sec:experimental_setup}

\paragraph{Models and Protocol.}
We evaluate 17 frozen MLLMs: 8 proprietary and 9 open-weight models
(Table~\ref{tab:foundation_main}), on the same 1,000 questions without
adaptation to the benchmark. A common input policy uses each model's
supported media interface: native video and native audio take priority,
with sampled frames (1 FPS, at most 64) and ASR as the respective
fallbacks. Memory questions include relevant history.
Input details and separately evaluated SFT/GRPO checkpoints appear in
Appendices~\ref{app:model_details} and~\ref{app:adaptation_analysis}.

\paragraph{Metrics and Scoring.}
Models produce an answer and structured rationales.
\textbf{Raw Answer} measures correctness; \textbf{Reasoning-Verified}
additionally considers rationale coverage, cue support, and logical integrity.
Both use macro averaging across eight capabilities.
Closed answers use deterministic matching; Qwen3.8-27B judges open answers
and rationales. Scoring and paired bootstrap procedures over episodes appear in
Appendix~\ref{app:evaluation_details}.

\subsection{Main Results}
\label{sec:foundation_evaluation}

\begin{table*}[!ht]
\centering
\caption{
\textbf{Main results on PsycheMME (\%).}
We report 17 frozen models and two adapted checkpoints.
Raw and Ver. are macro scores across eight capabilities;
capability columns report Verified scores.
Colors mark the top five scores in each column and model group.
Bold names mark the adapted Qwen3.5-4B checkpoints.
}
\label{tab:foundation_main}

\definecolor{psyYellow1}{HTML}{F2D8A7}
\definecolor{psyYellow2}{HTML}{F6E2BF}
\definecolor{psyYellow3}{HTML}{F9EBD5}
\definecolor{psyYellow4}{HTML}{FCF3E5}
\definecolor{psyYellow5}{HTML}{FEF9F2}
\definecolor{psyBlue1}{HTML}{BBD5EB}
\definecolor{psyBlue2}{HTML}{D0E2F2}
\definecolor{psyBlue3}{HTML}{E1EDF7}
\definecolor{psyBlue4}{HTML}{EDF4FA}
\definecolor{psyBlue5}{HTML}{F6FAFD}
{
\small
\renewcommand{\arraystretch}{1.06}
\setlength{\tabcolsep}{3.2pt}
\begin{tabular*}{\textwidth}{l@{\extracolsep{\fill}}cccccccccc}
\toprule[0.1em]
\multicolumn{1}{c}{
    \multirow{2}{*}{
        \raisebox{-0.3ex}{\textbf{Model}}
    }
}& \multicolumn{2}{c}{\textbf{Overall}}
& \multicolumn{3}{c}{\textbf{Understanding}}
& \multicolumn{3}{c}{\textbf{Interaction}}
& \multicolumn{2}{c}{\textbf{Contextual}} \\
\cmidrule(lr){2-3}\cmidrule(lr){4-6}\cmidrule(lr){7-9}\cmidrule(lr){10-11}
& Raw & Ver. & G & W & N & T & S & A & M & R \\
\midrule[0.055em]
\multicolumn{11}{c}{\rule[-0.55ex]{0pt}{2.8ex}\textbf{\textit{Proprietary Models}}} \\
\midrule[0.055em]
Gemini 3.6 Flash~\citep{googledeepmind2026gemini36flash}
& \cellcolor{psyBlue1}\textbf{85.68} & \cellcolor{psyYellow1}\textbf{80.03} & \cellcolor{psyBlue2}\underline{91.76} & \cellcolor{psyBlue1}\textbf{79.68} & \cellcolor{psyBlue1}\textbf{69.97} & \cellcolor{psyBlue1}\textbf{47.81} & \cellcolor{psyBlue2}\underline{79.32} & \cellcolor{psyBlue1}\textbf{89.58} & \cellcolor{psyBlue2}\underline{90.92} & \cellcolor{psyBlue1}\textbf{91.21} \\
Claude Opus 5~\citep{anthropic2026opus5}
& \cellcolor{psyBlue2}\underline{82.34} & \cellcolor{psyYellow2}\underline{76.71} & 89.31 & \cellcolor{psyBlue3}77.80 & \cellcolor{psyBlue3}69.10 & \cellcolor{psyBlue3}40.70 & \cellcolor{psyBlue3}79.18 & \cellcolor{psyBlue2}\underline{86.24} & \cellcolor{psyBlue4}86.56 & 84.82 \\
GPT-5.6 Sol~\citep{openai2026gpt56}
& \cellcolor{psyBlue5}81.09 & \cellcolor{psyYellow3}75.98 & \cellcolor{psyBlue1}\textbf{92.13} & \cellcolor{psyBlue2}\underline{78.75} & \cellcolor{psyBlue4}67.28 & 32.99 & \cellcolor{psyBlue4}74.86 & \cellcolor{psyBlue4}84.25 & \cellcolor{psyBlue1}\textbf{91.11} & \cellcolor{psyBlue5}86.47 \\
Qwen3.8-Max~\citep{qwenteam2026qwen38}
& \cellcolor{psyBlue3}81.48 & \cellcolor{psyYellow4}75.59 & \cellcolor{psyBlue5}90.87 & \cellcolor{psyBlue4}77.61 & \cellcolor{psyBlue2}\underline{69.17} & \cellcolor{psyBlue4}39.33 & \cellcolor{psyBlue5}72.84 & \cellcolor{psyBlue5}82.46 & 84.48 & \cellcolor{psyBlue2}\underline{87.94} \\
Kimi K3~\citep{kimitteam2026kimi3}
& 80.60 & \cellcolor{psyYellow5}75.35 & \cellcolor{psyBlue3}91.52 & 76.11 & 60.62 & \cellcolor{psyBlue5}38.82 & \cellcolor{psyBlue1}\textbf{79.38} & \cellcolor{psyBlue3}84.28 & \cellcolor{psyBlue5}85.17 & \cellcolor{psyBlue3}86.87 \\
Gemini 3.1 Pro~\citep{googledeepmind2026gemini31pro}
& \cellcolor{psyBlue4}81.36 & 74.33 & \cellcolor{psyBlue4}91.30 & 76.26 & 60.29 & \cellcolor{psyBlue2}\underline{44.71} & 68.22 & 76.80 & \cellcolor{psyBlue3}90.49 & \cellcolor{psyBlue4}86.57 \\
GLM-5V-Turbo~\citep{glmVteam2026glm5vturbo}
& 78.40 & 71.19 & 89.18 & 76.72 & \cellcolor{psyBlue5}61.82 & 34.23 & 68.14 & 79.87 & 76.23 & 83.37 \\
Seed 2.1 Pro~\citep{seedteam2026seed21}
& 76.37 & 69.67 & 88.53 & \cellcolor{psyBlue5}76.87 & 53.24 & 32.93 & 63.27 & 78.09 & 82.42 & 81.98 \\
\specialrule{0.075em}{0.16em}{0.04em}
\multicolumn{11}{c}{\rule[-0.55ex]{0pt}{2.8ex}\textbf{\textit{Open-weight Models}}} \\
\specialrule{0.055em}{0.10em}{0pt}
\rowcolor{black!10}
\multicolumn{11}{l}{\rule[-0.55ex]{0pt}{2.8ex}\bfseries\itshape\boldmath Model Size $> 8$B} \\
Qwen3.8-27B~\citep{qwen38}
& \cellcolor{psyBlue1}\textbf{79.97} & \cellcolor{psyYellow1}\textbf{72.74} & \cellcolor{psyBlue1}\textbf{90.93} & \cellcolor{psyBlue1}\textbf{80.53} & \cellcolor{psyBlue1}\textbf{66.96} & \cellcolor{psyBlue2}\underline{28.58} & \cellcolor{psyBlue1}\textbf{68.87} & \cellcolor{psyBlue1}\textbf{83.01} & \cellcolor{psyBlue1}\textbf{77.77} & \cellcolor{psyBlue1}\textbf{85.27} \\
Qwen3.5-9B~\citep{qwenteam2026qwen35}
& \cellcolor{psyBlue2}\underline{76.54} & \cellcolor{psyYellow2}\underline{66.45} & \cellcolor{psyBlue2}\underline{90.24} & \cellcolor{psyBlue2}\underline{76.86} & \cellcolor{psyBlue2}\underline{54.66} & \cellcolor{psyBlue1}\textbf{29.13} & \cellcolor{psyBlue2}\underline{62.63} & \cellcolor{psyBlue2}\underline{73.45} & \cellcolor{psyBlue3}68.14 & \cellcolor{psyBlue3}76.52 \\
Gemma-4-12B-it~\citep{gemmateam2026gemma4}
& \cellcolor{psyBlue5}66.41 & \cellcolor{psyYellow4}57.75 & \cellcolor{psyBlue4}79.48 & \cellcolor{psyBlue3}71.19 & \cellcolor{psyBlue3}46.41 & 13.49 & \cellcolor{psyBlue5}49.43 & \cellcolor{psyBlue4}62.32 & \cellcolor{psyBlue5}65.07 & \cellcolor{psyBlue4}74.58 \\
InternVL3.5-14B~\citep{wang2025internvl35}
& 52.24 & 44.68 & 63.36 & 56.63 & 19.90 & 15.63 & 34.05 & 45.10 & 59.53 & 63.22 \\
\specialrule{0.055em}{0.10em}{0pt}
\rowcolor{black!10}
\multicolumn{11}{l}{\rule[-0.55ex]{0pt}{2.8ex}\bfseries\itshape\boldmath $4$B $<$ Model Size $\leq 8$B} \\
MiniCPM-V4.5~\citep{openbmb2025minicpm}
& \cellcolor{psyBlue3}69.65 & \cellcolor{psyYellow3}58.13 & \cellcolor{psyBlue3}81.26 & \cellcolor{psyBlue4}70.63 & 33.81 & \cellcolor{psyBlue3}24.32 & \cellcolor{psyBlue4}50.08 & \cellcolor{psyBlue3}62.45 & \cellcolor{psyBlue4}65.40 & \cellcolor{psyBlue2}\underline{77.13} \\
InternVL3.5-8B~\citep{wang2025internvl35}
& \cellcolor{psyBlue4}67.41 & \cellcolor{psyYellow5}55.41 & 72.30 & \cellcolor{psyBlue5}64.39 & \cellcolor{psyBlue4}35.38 & 20.65 & \cellcolor{psyBlue3}50.37 & \cellcolor{psyBlue5}59.32 & \cellcolor{psyBlue2}\underline{72.20} & 68.67 \\
Phi-4-multimodal~\citep{abouelenin2025phi4multimodal}
& 61.03 & 48.63 & 52.50 & 50.32 & 22.14 & 20.85 & 45.96 & 58.17 & 64.63 & \cellcolor{psyBlue5}74.44 \\
\specialrule{0.055em}{0.10em}{0pt}
\rowcolor{black!10}
\multicolumn{11}{l}{\rule[-0.55ex]{0pt}{2.8ex}\bfseries\itshape\boldmath Model Size $\leq 4$B} \\
Qwen3.5-4B \textbf{(SFT+GRPO)}
& 73.19 & 52.62 & 83.43 & 67.75 & 37.64 & 17.24 & 47.68 & 60.82 & 46.53 & 59.86 \\
Qwen3.5-4B \textbf{(SFT)}
& 72.58 & 52.13 & 82.73 & 67.83 & 36.92 & 15.16 & 48.02 & 59.38 & 47.45 & 59.59 \\
InternVL3.5-4B~\citep{wang2025internvl35}
& 61.30 & 48.97 & \cellcolor{psyBlue5}72.46 & 60.06 & 26.90 & \cellcolor{psyBlue5}21.61 & 37.44 & 47.00 & 63.74 & 62.54 \\
Qwen3.5-4B~\citep{qwenteam2026qwen35}
& 47.68 & 37.71 & 60.70 & 46.49 & \cellcolor{psyBlue5}34.18 & \cellcolor{psyBlue4}23.17 & 24.02 & 35.85 & 33.61 & 43.66 \\
\bottomrule[0.1em]
\end{tabular*}
}
\end{table*}

\begin{wrapfigure}{R}{0.47\columnwidth}
    \centering
    \includegraphics[width=0.98\linewidth]{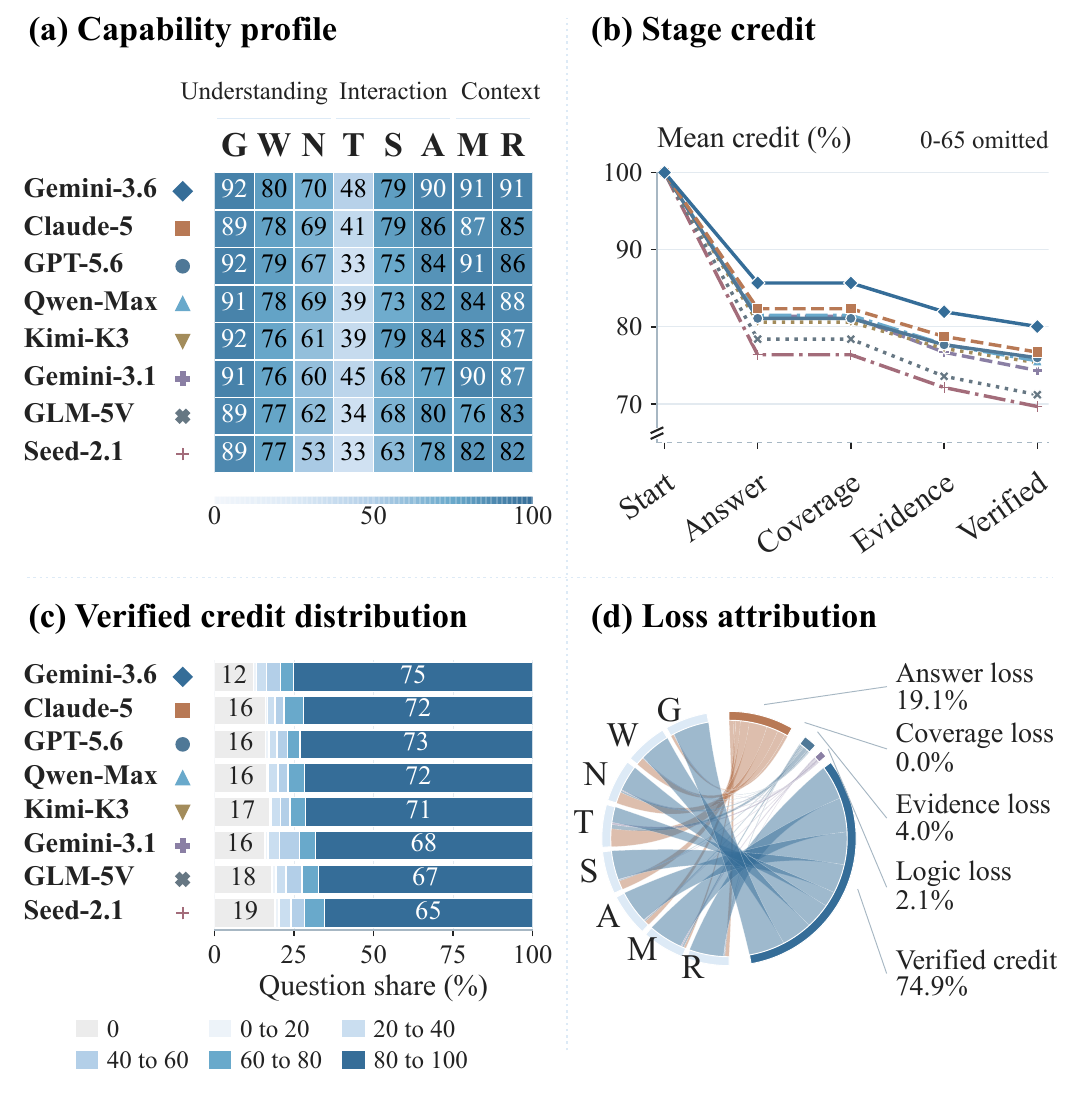}
    \vspace{-12pt}
    \caption{
    \textbf{Performance breakdown.}
Performance by capability, verification stages, score distributions, and loss attribution.
    }
    \label{fig:task_progress}
\end{wrapfigure}

Table~\ref{tab:foundation_main} shows that Gemini 3.6 Flash leads overall
(85.68/80.03 Raw/Verified), while Qwen3.8-27B leads open-weight models
(79.97/72.74). Figure~\ref{fig:task_progress} summarizes capability profiles
and verification losses.

\noindent\textbf{(1) Timing is difficult to answer and justify.}
Timing has the lowest Verified score and credit retention (Verified/Raw)
for all 17 models; the retention pattern persists on single-choice questions.
Even among 569 correct Timing responses, 283 receive weak or contradicted
Evidence Support labels (49.74\%), compared with 7.62\% of other correct
single-choice responses. The difference is 42.12 percentage points
(95\% CI $[33.37,51.50]$; Appendix~\ref{app:foundation_audit}).

\begin{figure*}[!b]
    \centering
    \includegraphics[width=\textwidth]{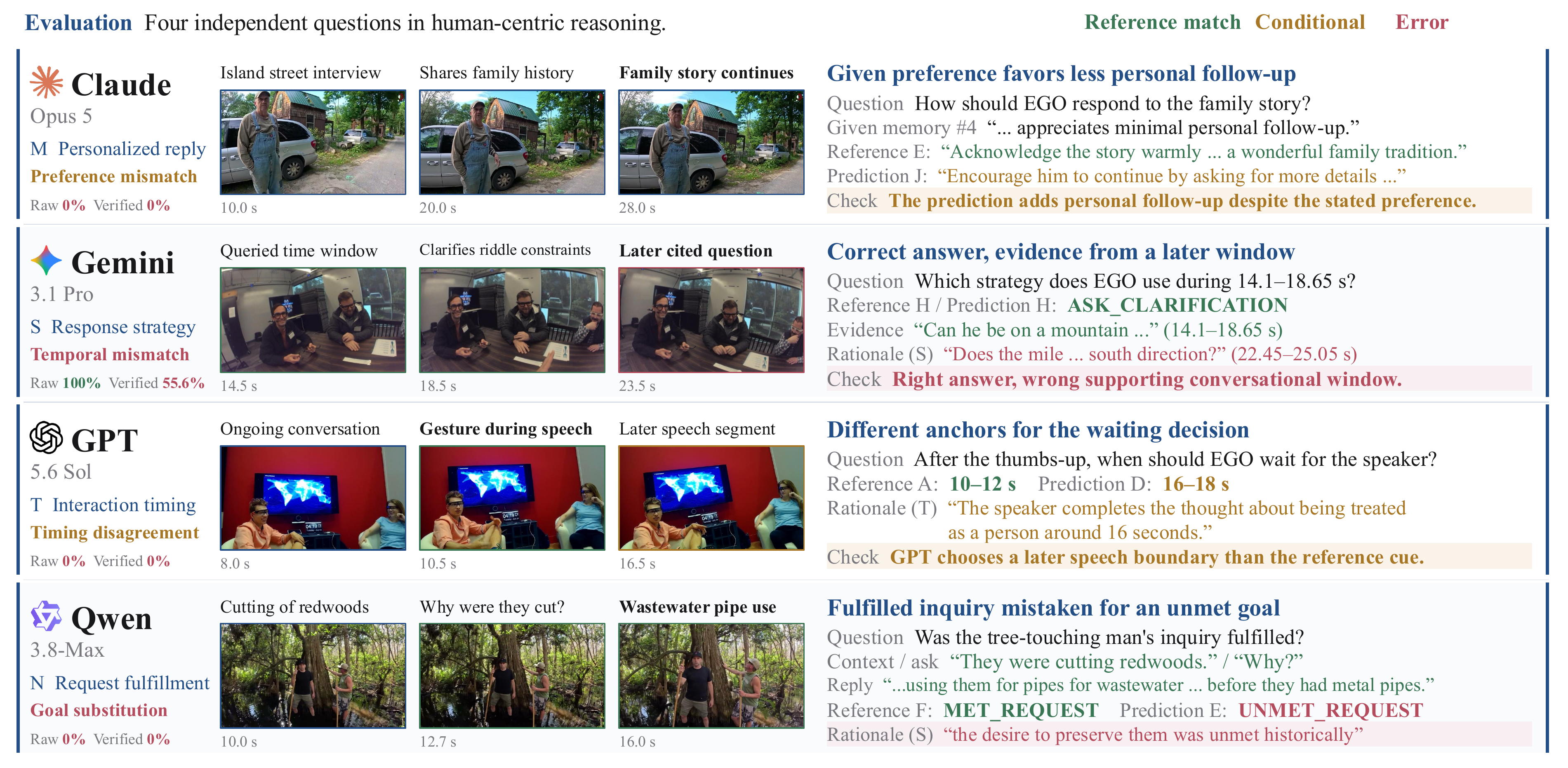}
    \caption{
    \textbf{Illustrative model responses in PsycheMME.}
    The cases illustrate preference mismatch, temporal cue
    mismatch, disagreement about interaction timing, and goal substitution.
    }
    \label{fig:failure_case}
\end{figure*}

\noindent\textbf{(2) Similar Need scores accompany different answer completeness.}
Gemini 3.6 Flash and Claude Opus 5 have similar Need Verified scores
(69.97/69.10). Flash leads on choice questions (81.31/70.95 Verified),
whereas Claude leads on open questions (78.57/57.14 Raw).

On the 35 open Need questions, Flash and Claude receive full credit on
28.57/68.57\% of items, despite similar rates of nonzero credit
(85.71/88.57\%). This difference of 40.00 percentage points in full credit
accounts for 20.00 of their 21.43 point Raw gap under the $0/0.5/1$ rubric.
Across all 110 open questions, 38 receive full answer credit from both
models; Flash has higher mean Verified credit on these questions
(97.15 versus 90.42). Answer completeness and rationale quality thus favor
different models (Appendix~\ref{app:foundation_audit}).

\begin{table*}[!t]
\centering
\begin{minipage}[t]{0.48\textwidth}
    \vspace{0pt}
    \centering

\definecolor{ablBlue1}{HTML}{BBD5EB}
\definecolor{ablBlue2}{HTML}{D0E2F2}
\definecolor{ablBlue3}{HTML}{E1EDF7}

\definecolor{ablYellow1}{HTML}{F2D8A7}
\definecolor{ablYellow2}{HTML}{F6E2BF}
\definecolor{ablYellow3}{HTML}{F9EBD5}

\definecolor{ablGray}{HTML}{F0F0F0}
\definecolor{ablGrayText}{HTML}{73777B}
\definecolor{ablGreen}{HTML}{388457}

\captionof{table}{
\textbf{Ablation studies on PsycheMME (\%).}
Green indicates gains over Question-only;
$\Delta$ denotes gains over baselines.
(Appendix~\ref{app:ablation}).}

\label{tab:modality}

\label{tab:app_ablation}

\begingroup
\footnotesize
\setlength{\tabcolsep}{3pt}
\renewcommand{\arraystretch}{1.12}

\newcommand{\abGain}[1]{%
  \textsuperscript{%
    \textcolor{ablGreen}{\tiny +#1}%
  }%
}

\newcommand{\abDelta}[1]{%
  \textcolor{ablGreen}{#1}%
}

\resizebox{0.98\linewidth}{!}{%
\begin{tabular}{@{}lcccc@{}}

\toprule[0.09em]

& \multicolumn{2}{c}{\textbf{Qwen3.5-9B}}
& \multicolumn{2}{c}{\textbf{MiniCPM-V4.5}} \\

\cmidrule(lr){2-3}
\cmidrule(l){4-5}

\textbf{Input}
& Raw & Ver. & Raw & Ver. \\

\midrule

\rowcolor{ablGray}
\multicolumn{5}{@{}l@{}}{
  \textcolor{ablGrayText}{
    \textit{Modality controls}
  }
} \\

Question-only
& 64.04
& 57.05
& 57.22
& \cellcolor{ablYellow3}49.93 \\

ASR-only
& \cellcolor{ablBlue3}67.04
& \cellcolor{ablYellow3}57.75
& \cellcolor{ablBlue3}60.05
& 49.36 \\

Visual-only
& \cellcolor{ablBlue2}\underline{73.44}
& \cellcolor{ablYellow2}\underline{65.50}
& \cellcolor{ablBlue2}\underline{65.67}
& \cellcolor{ablYellow2}\underline{55.56} \\

\textbf{Full}
& \cellcolor{ablBlue1}\textbf{76.54}\abGain{12.50}
& \cellcolor{ablYellow1}\textbf{66.45}\abGain{9.40}
& \cellcolor{ablBlue1}\textbf{69.65}\abGain{12.43}
& \cellcolor{ablYellow1}\textbf{58.13}\abGain{8.20} \\

\textit{$\Delta$ vs. ASR-only}
& \abDelta{+9.50}
& \abDelta{+8.70}
& \abDelta{+9.60}
& \abDelta{+8.77} \\

\textit{$\Delta$ vs. Visual-only}
& \abDelta{+3.10}
& \abDelta{+0.95}
& \abDelta{+3.98}
& \abDelta{+2.57} \\

\addlinespace[3pt]
\midrule[0.04em]

\rowcolor{ablGray}
\multicolumn{5}{@{}l@{}}{
  \textcolor{ablGrayText}{
    \textit{Frame control}
  }
} \\

Single-image
& 72.28 & 61.78
& 66.16 & 54.99 \\

\bottomrule[0.09em]

\end{tabular}%
}

\endgroup

\end{minipage}\hfill
\begin{minipage}[t]{0.48\textwidth}
    \vspace{0pt}
    \centering
\captionsetup{skip=5pt,singlelinecheck=false}
\definecolor{memGreen}{HTML}{388457}
\captionof{table}{
\textbf{Memory intervention (\%).}
Green superscripts show gains over No History (points);
Mismatched History averages three seeds.
}
\label{tab:memory_intervention}

\small
\renewcommand{\arraystretch}{1.12}
\setlength{\tabcolsep}{6pt}

\begin{tabular}{p{\dimexpr0.44\linewidth-2\tabcolsep\relax}
    >{\centering\arraybackslash}p{\dimexpr0.28\linewidth-2\tabcolsep\relax}
    >{\centering\arraybackslash}p{\dimexpr0.28\linewidth-2\tabcolsep\relax}}
    \toprule
    \textbf{History Condition}
    & \textbf{Raw} $\uparrow$
    & \textbf{Ver.} $\uparrow$ \\
    \midrule

    No History
    & 42.50 & 25.28 \\

    Mismatched History
    & 45.00 & 23.87 \\

    \rowcolor{aprilblue!12}
    \textbf{Matched History}
    & \textbf{82.50}$^{\scriptscriptstyle \textcolor{memGreen}{+40.00}}$
    & \textbf{74.55}$^{\scriptscriptstyle \textcolor{memGreen}{+49.27}}$ \\

    \bottomrule
\end{tabular}

    \par\medskip
    \centering
\captionsetup{skip=5pt,singlelinecheck=false}

\captionof{table}{
\textbf{Human baseline (\%).}
Five annotators independently answer 300 randomly sampled questions.
}
\label{tab:human_baseline}

\small
\renewcommand{\arraystretch}{1.12}
\setlength{\tabcolsep}{6pt}

\begin{tabular*}{\linewidth}{l@{\extracolsep{\fill}}ccc}
    \toprule
    \textbf{Annotators}
    & \textbf{Questions}
    & \textbf{Micro} $\uparrow$
    & \textbf{Macro} $\uparrow$ \\
    \midrule

    5
    & 300
    & 85.27
    & 84.82 \\

    \bottomrule
\end{tabular*}

\end{minipage}
\end{table*}

\noindent\textbf{(3) Complementary answers do not ensure reliable consensus.}
On Timing, the eight proprietary models offer 91.67\% oracle coverage,
compared with 66.67\% accuracy for the best model; plurality voting reaches
only 57.64\% with uniform tie breaking.
At least five models select the same wrong option on 21 questions;
18 nevertheless contain a correct candidate.
Shared wrong choices can outvote correct minority candidates.
Voting details and paired intervals appear in
Appendix~\ref{app:failure_analysis}.

In Figure~\ref{fig:failure_case}, Gemini uses cues from the wrong time
window to justify a correct strategy; GPT selects a later intervention
boundary. Claude misses a stated preference; Qwen substitutes a different goal.

\subsection{Ablations and Validity Analysis}
\label{sec:further_analysis}

\paragraph{Multimodal Cues.}
Full video and ASR improve Raw scores over Question-only by
12.50/12.43 points for Qwen3.5-9B/MiniCPM-V4.5
(Table~\ref{tab:modality}). Visual-only retains most Full accuracy;
adding ASR gives another 3.10/3.98 points. Removing video (ASR-only)
reduces Grounding by 30.65/45.16 Raw points. Using a single midpoint
image reduces Qwen's Grounding by 12.63 points and Timing by 8.33 points
(Appendix~\ref{app:modality_analysis}).

\paragraph{Historical Context.}
On 40 Memory questions, matched history raises Qwen3.5-9B Raw/Verified
from 42.50/25.28 to 82.50/74.55, while mismatched history averages
45.00/23.87 (Table~\ref{tab:memory_intervention}).
The gains depend on relevant history; results for each seed
appear in Appendix~\ref{app:memory_analysis}.

\paragraph{Task Adaptation.}
Qwen3.5-4B improves from 47.68/37.71 Raw/Verified macro scores to
72.58/52.13 after SFT and 73.19/52.62 after SFT+GRPO
(Table~\ref{tab:foundation_main}). Gains vary by capability: SFT raises
Grounding and Risk Raw scores by 27.96/27.40 points, but Timing falls
from 40.28 to 31.94 and recovers only to 34.72 after GRPO
(Table~\ref{tab:psychebench-sft-grpo}). GRPO adds 0.61 Raw and 0.49
Verified points, leaving a gap of 20.57 points between Raw and Verified.
These results leave temporal grounding and rationale quality unresolved.

\paragraph{Human Baseline.}
Five annotators independently answer 300 randomly sampled questions
(1,500 responses), achieving 85.27\% Raw micro and 84.82\% macro accuracy
(Table~\ref{tab:human_baseline}). Timing accuracy is 79.43\% on this
sample (Appendix~\ref{app:human_evaluation}).

\paragraph{Human-Judge Agreement.}
Five psychology experts independently review 300 randomly sampled Judge cases:
1,352 of 1,500 assessments endorse the presented decisions
(90.13\%; Appendix~\ref{app:human_endorsement}).
Weight sensitivity analysis supports the Timing retention
finding (Appendix~\ref{app:weight_sensitivity}).

\section{Conclusion}
\label{sec:conclusion}

PsycheMME unifies eight capabilities informed by psychology in a
benchmark of 1,000 questions grounded in video. Its evaluation protocol assesses
answers alongside support from reference cues and the logical integrity
of explicit rationales. Across 17 models, Timing remains difficult,
similar Need scores accompany differences in answer completeness, and voting
can override correct minority answers. Input and history controls
reveal contextual dependence.
These findings describe behavior by capability and justification
quality beyond aggregate accuracy.
The resulting diagnostics support comparisons of human understanding and
interaction judgments while exposing how answer completeness and rationale
support shape reported performance.
The benchmark can guide future improvements in answer completeness,
temporal grounding, and response selection informed by context.

\section*{Limitations}
PsycheMME evaluates human-centered reasoning
in recorded social interactions; real-time
interactive settings remain unevaluated. While the benchmark
covers diverse scenarios, its evaluation remains
bounded by the represented social and cultural
contexts. Future work could expand contextual and
cultural diversity, explore streaming and interactive
evaluation, and incorporate more fine-grained human
assessments of explanation quality.

\clearpage
\bibliography{custom}

\clearpage
\appendix
\raggedbottom
\renewcommand{\topfraction}{0.9}
\renewcommand{\dbltopfraction}{0.9}
\renewcommand{\textfraction}{0.08}
\renewcommand{\floatpagefraction}{0.75}
\renewcommand{\dblfloatpagefraction}{0.75}
\setcounter{topnumber}{3}
\setcounter{dbltopnumber}{3}
\setcounter{totalnumber}{5}
\setlength{\textfloatsep}{10pt plus 2pt minus 2pt}
\setlength{\dbltextfloatsep}{6pt plus 2pt minus 1pt}
\setlength{\floatsep}{8pt plus 2pt minus 2pt}
\setlength{\dblfloatsep}{5pt plus 2pt minus 1pt}
\setlength{\intextsep}{8pt plus 2pt minus 2pt}
\setlength{\abovecaptionskip}{6pt}
\providecommand{\PsycheExpRoot}{.}
\providecommand{\appfill}{\rule{0.65cm}{0.35pt}}
\newsavebox{\psycheapptablebox}
\newcommand{\psycheapptablefit}[1]{%
  \sbox{\psycheapptablebox}{#1}%
  \ifdim\wd\psycheapptablebox>\linewidth
    \resizebox{\linewidth}{!}{\usebox{\psycheapptablebox}}%
  \else
    \usebox{\psycheapptablebox}%
  \fi}
\providecommand{\psycheappfigure}[3]{%
  \begin{figure*}[!t]
  \centering
  \includegraphics[width=\textwidth]{\PsycheExpRoot/figures/appendix/#1.pdf}
  \caption{#3}\label{#2}
  \end{figure*}}

\begingroup
\setlength{\parskip}{0pt}
\section*{Appendix Contents}
\fontsize{9.5}{11.5}\selectfont
\setlength{\tabcolsep}{5pt}
\renewcommand{\arraystretch}{1}
\newcommand{\appcontentssection}[2]{%
  \addlinespace[4pt]
  \textbf{\textcolor{psyBlueText}{\ref{#1}\quad\hyperref[#1]{#2}}}
  & & \textcolor{psyBlueText}{\pageref{#1}}\\}
\newcommand{\appcontentsevidence}[3]{%
  \hspace{1.5em}\textcolor{psyBlueText}{\hyperref[#2]{#3}}
  & \textcolor{psyBlueText}{\hyperref[#2]{#1~\ref{#2}}}
  & \textcolor{psyBlueText}{\pageref{#2}}\\}
\noindent\begin{tabular*}{\textwidth}{@{}l@{\extracolsep{\fill}}lr@{}}
\toprule
Section / item & Reference & Page\\
\midrule
\appcontentssection{app:related_work}{Benchmark Comparison Criteria}
\appcontentssection{app:taxonomy_details}{Psychological Foundations and Capability Definitions}
\appcontentssection{app:construction_pipeline}{Benchmark Construction and Quality Review}
\appcontentsevidence{Figure}{fig:app_construction_workflow}{Question construction and review}
\appcontentsevidence{Table}{tab:app_source_composition}{Benchmark composition by source}
\appcontentsevidence{Table}{tab:app_denominators}{Capability and question format counts}
\appcontentsevidence{Table}{tab:app_agent_access}{Construction agent roles}
\appcontentsevidence{Table}{tab:app_review_codebook}{Human question review criteria}
\appcontentssection{app:experimental_details}{Experimental and Evaluation Protocol}
\appcontentsevidence{Figure}{fig:app_scoring_workflow}{Evaluation and scoring procedure}
\appcontentsevidence{Table}{tab:app_response_schema}{Response fields and answer representations}
\appcontentsevidence{Table}{tab:app_judge_rubric}{Rationale labels and scoring multipliers}
\appcontentsevidence{Table}{tab:app_scoring_examples}{Examples of Verified credit}
\appcontentssection{app:post_training_details}{SFT and GRPO on PsycheMME}
\appcontentsevidence{Figure}{fig:app_training_comparison}{SFT and GRPO comparison}
\appcontentsevidence{Table}{tab:psychebench-sft-grpo}{Base, SFT, and GRPO results}
\appcontentssection{app:foundation_statistics}{Complete Foundation Model Results}
\appcontentsevidence{Figure}{fig:app_capability_heatmap}{Verified capability profiles}
\appcontentsevidence{Table}{tab:app_raw}{Raw scores by capability with confidence intervals}
\appcontentsevidence{Table}{tab:app_verified}{Verified scores by capability with confidence intervals}
\appcontentsevidence{Table}{tab:app_main_uncertainty}{Overall confidence intervals}
\appcontentssection{app:foundation_audit}{Conditional Rationale and Stratified Analysis}
\appcontentsevidence{Table}{tab:app_conditional_support}{Verification of correct single-choice answers}
\appcontentsevidence{Table}{tab:app_need_formats}{Need scores and open answer completeness}
\appcontentsevidence{Table}{tab:app_open_diagnostics}{Open answer credit and answer length}
\appcontentsevidence{Table}{tab:app_scoring_sensitivity}{Scoring weights and aggregation sensitivity}
\appcontentsevidence{Table}{tab:app_formats}{Scores by question format}
\appcontentsevidence{Table}{tab:app_sources}{Scores by source}
\appcontentsevidence{Table}{tab:app_strata}{Scores by duration and question subset}
\appcontentssection{app:failure_analysis}{Complementary Answers and Voting}
\appcontentsevidence{Table}{tab:app_closed_shared}{Answer overlap and plurality voting}
\appcontentssection{app:modality_analysis}{Multimodal Cues and Input Controls}
\appcontentsevidence{Figure}{fig:app_modality_comparison}{Modality and single image controls}
\appcontentsevidence{Table}{tab:app_modality}{Modality ablations}
\appcontentssection{app:memory_analysis}{Historical Context and Memory Controls}
\appcontentsevidence{Figure}{fig:app_memory_context}{History context controls}
\appcontentsevidence{Table}{tab:app_memory}{History interventions}
\appcontentssection{app:human_evaluation}{Human Baseline}
\appcontentsevidence{Table}{tab:app_human_results}{Human answer accuracy by annotator}
\appcontentsevidence{Table}{tab:app_human_capabilities}{Human answer accuracy by capability}
\appcontentssection{app:human_endorsement}{Human-Judge Agreement}
\appcontentsevidence{Table}{tab:app_human_judge_final}{Expert endorsement of Judge decisions}
\appcontentsevidence{Table}{tab:app_human_judge_capabilities}{Human-Judge Agreement by capability}
\appcontentssection{app:reproducibility}{Analysis Scope and Reproducibility}
\appcontentsevidence{Table}{tab:app_analysis_scopes}{Sample sizes and evaluation units}
\bottomrule
\end{tabular*}
\endgroup
\clearpage

\providecommand{\aprpartial}{\ensuremath{\triangle}}

\section{Benchmark Comparison Criteria}
\label{app:related_work}
\label{app:benchmark_comparison}

Table~\ref{tab:benchmark_comparison} compares
the evaluation settings reported by existing
multimodal benchmarks of human understanding and interaction.
We distinguish three levels of coverage:
\aprcheck\ indicates explicit evaluation
through dedicated tasks or scoring criteria;
\aprpartial\ indicates restricted evaluation
through specific subtasks or components;
and \aprx\ indicates a reported setting without
dedicated evaluation of that capability.

The eight dimensions follow the capability
taxonomy defined in our benchmark.
Coverage requires a task or scoring criterion
that addresses the corresponding target judgment.
Interaction timing concerns when to respond;
embodied action selection concerns which action to take;
and memory evaluation requires judgments that use
relevant historical information.

HumanSense receives restricted Memory coverage because its
Psychological Chat task conditions response selection on prior
multiple dialogue turns~\cite{qin2026humansense}.
SocialReasonBench's task on moral dilemmas predicts majority player
choices rather than normative appropriateness, so it does not
establish Risk coverage~\cite{huang2026socialreasonbench}.
Human Behavior Atlas evaluates affect, intent, social reasoning,
and nonverbal communication. We mark Grounding as restricted to
interpreting behavioral cues; its tasks do not require a unified
assessment of intervention timing, action selection, or boundaries
\cite{ong2026human}.

We assign coverage based on the reported
task definitions and evaluation protocols.
The Ego column indicates the use of
egocentric observations.
Videos/Clips and Prompt counts retain the units
reported by the respective papers and are
interpreted within each benchmark's source and task scope.

\section{Psychological Foundations and Capability Definitions}
\label{app:taxonomy_details}

This section expands the eight capabilities in
Sec.~\ref{sec:capability_taxonomy}, specifying their theoretical
foundations, target judgments, and requirements for observed cues.
Schematic interaction examples illustrate the distinctions between
these judgments.

\subsection{Theoretical Foundations and the Eight Capabilities}
\label{app:taxonomy_organization}

The taxonomy connects social observation to contextual response
selection. Social Information Processing distinguishes cue encoding and
interpretation from goal clarification and response
selection~\cite{crick1994sip}; the Perception-Action Cycle describes
how perception and temporally organized information inform
action~\cite{fuster2004perception}.
We operationalize these principles through judgments with explicit
targets and requirements for observed cues.

\emph{Perceptually Grounded Mentalizing} distinguishes observable cues,
psychological interpretations, and support needs.
\emph{Situated Empathic Responding} distinguishes intervention timing,
verbal response, and physical or nonverbal action, informed by empathy
and interpersonal emotion regulation~\cite{davis1983empathy,zaki2013interpersonal}.
\emph{Contextual and Normative Modulation} addresses historical
information and constraints affecting either domain.

The eight capabilities separate six judgment targets and two forms of
contextual modulation. G, W, and N address what is observed, how the
person's state is interpreted, and what support is needed; T, S, and A
address when to respond, what to say, and what action to take.
M and R specify how relevant history and consent or safety constraints
shape those judgments. A question's requested judgment and contextual
requirements determine its capability assignment, enabling diagnosis
at each of these points in social understanding and response selection.
G/W/N/T/S/A/M/R denote Grounding, Understanding, Need, Timing,
Response, Action, Memory, and Risk throughout.

\subsection{Capability Definitions}
\label{app:taxonomy_definitions}

\subsubsection{Perceptually Grounded Mentalizing}

This domain relates observable interaction structure to human states
and needs through cues about each person and the supplied context.

\paragraph{Social Grounding (G).}
Grounding identifies participants, observable behavior, and interaction
relations: who speaks, who is addressed, and whose gesture or action
is described. Speech, expression, gaze, and movement provide
cues specific to each person; turn structure and spatial context distinguish
speakers, addressees, and targets. Associating a cue with the correct
participant establishes what was observed. Explaining the person's
underlying state instead belongs to Understanding.

\paragraph{Psychological Understanding (W).}
Understanding concerns psychological states and their possible causes
or changes during an interaction. Theory of Mind motivates the
distinction between observed behavior and attributed mental
states~\cite{premack1978tom}. Judgments integrate dialogue, behavior,
and context to interpret converging or conflicting cues.
The target is an interpretation of the person's state supported
by observed cues within the supplied interaction.

\paragraph{Need Inference (N).}
Need identifies what a person wants to change, what support may be
relevant, and whether they would accept it. Requests, hesitation,
refusal, repeated difficulty, and affective reactions inform these
judgments. Distinguishing practical help, acknowledgment, or space requires
cues indicating what support the person wants. N thus distinguishes possible
support needs from willingness to receive help; T, S, and A address
when and how to respond.

\subsubsection{Situated Empathic Responding}

This domain evaluates intervention opportunity, verbal content, and
behavioral form as distinct aspects of response appropriateness.

\paragraph{Interaction Timing and Attunement (T).}
Timing judges whether to respond now, continue listening, wait,
or refrain. Conversational turn-taking motivates attention to ongoing
speech and participation~\cite{sacks1974turntaking}; PsycheMME also
considers activity, urgency, pauses, and changes across the interaction.
Turn structure and current activity establish whether a pause offers
an appropriate intervention opportunity.

\paragraph{Empathic Verbal Responding (S).}
Response selects what to say and how to say it when a verbal
intervention is appropriate. Content, strategy, and tone should fit
the person's state, need, and context, consistent with research on
matching support to stressful circumstances~\cite{cutrona1992support}.
Suitable responses include acknowledgment, validation, clarification,
advice, or a restrained reply. S evaluates the fit of response content,
strategy, and tone to the situation, complementing N's support need
and T's intervention opportunity.

\paragraph{Embodied Social Action (A).}
Action selects appropriate physical or nonverbal behavior, including
gaze, gesture, interpersonal distance, assistance, approach, retreat,
and handover. Judgments combine the person's state and need with
physical and social context. Recognizing an action already performed
belongs to G; choosing a subsequent action belongs to A.
The evaluation target is the appropriateness of the behavioral choice.

\subsubsection{Contextual and Normative Modulation}

Memory provides continuity across interactions; Risk constrains
otherwise feasible or helpful responses.

\paragraph{Interaction and Profile Memory (M).}
Memory uses relevant supplied history to interpret the current person
and situation. \emph{Interaction memory} includes prior events,
utterances, requests, refusals, and outcomes; \emph{profile memory}
includes individual preferences, habits, boundaries, and choices,
connecting to personalized multimodal reasoning~\cite{nie2026personavlm}.
Available history establishes individual preferences and
constraints. These categories organize the historical information
supplied to Memory questions.

\paragraph{Risk, Boundary, and Autonomy Regulation (R).}
Risk addresses harm, consent, personal boundaries, and autonomy.
Embodied evaluation focused on safety motivates attention to action
consequences~\cite{ying2026agentsafe}; PsycheMME also considers
expressed preferences and refusals. Constraints may require changing,
delaying, or abandoning an intervention. N identifies support needs; R checks the proposed intervention
against consent and other relevant constraints.

\subsection{Examples of Capability Distinctions}
\label{app:taxonomy_examples}

\paragraph{Observation, state, need, and timing.}
Someone frowns while attempting a task and says,
\emph{Let me finish explaining before you help}.
G identifies the actor and speaker; W interprets the supported state;
N identifies support needs; T favors listening before intervening.

\paragraph{Recognition and response selection.}
For an offered object, identifying the giver and recipient targets G;
choosing an acknowledgment targets S; deciding whether to accept or
step back targets A. Response choices also depend on requests and refusals.

\paragraph{History and present constraints.}
If history records a preference for verbal guidance, M applies it
while S and A select a fitting response. A present refusal invokes R,
making current consent decisive for the proposed intervention.

\subsection{Cue Grounding and Diagnostic Interpretation}
\label{app:taxonomy_evidence}

Expressions, utterances, and actions may support multiple psychological
accounts. Valid interpretations fit the supplied situation; responses
require cues supporting their timing, content, and form. Requests and
refusals establish willingness to receive support, while relevant
history establishes individual preferences.

The capability definitions are implemented through questions with
specified targets, reference cues, and explicit review criteria.
Five researchers with psychology backgrounds independently review
question validity, reference correctness, and psychological consistency;
benchmark inclusion requires unanimous approval
(Appendix~\ref{app:quality_details}).
Table~\ref{tab:app_review_codebook} specifies the cue requirements and
review criteria for each capability.

Capability profiles summarize performance on these target judgments
under the shared scoring protocol in
Sec.~\ref{sec:evaluation_protocol} and
Appendix~\ref{app:evaluation_details}.
They locate differences in observation, psychological interpretation,
support inference, response selection, use of history, and respect for
consent and safety constraints.

\section{Benchmark Construction and Quality Review}
\label{app:construction_pipeline}
\subsection{Source Composition and Cue Preparation}
\label{app:composition_details}

PsycheMME contains 630 video episodes from eight source groups
(Table~\ref{tab:app_source_composition}). An episode may support several
questions, yielding 1,000 questions in total: 890 choice questions
and 110 open questions. These statistics describe the
final benchmark after screening and quality review. Capability counts and
question format distributions appear in Table~\ref{tab:app_denominators}.

\begin{table*}[!t]
\centering
\caption{Final benchmark composition by source. Episodes are distinct within each source; Choice and Open denote choice and open questions. Counts sum to 630 episodes and 1,000 questions.}
\label{tab:app_source_composition}
\small
\setlength{\tabcolsep}{8pt}
\renewcommand{\arraystretch}{1.10}
\begin{tabular}{lrrrr}
\toprule
Source & Episodes & Questions & Choice & Open \\
\midrule
CASTLE & 61 & 87 & 84 & 3  \\
EasyCom & 40 & 70 & 66 & 4  \\
Ego4D & 142 & 230 & 196 & 34  \\
EgoCom & 38 & 60 & 48 & 12  \\
IndEgo & 8 & 8 & 8 & 0  \\
MERCI+ & 67 & 91 & 66 & 25  \\
YouTube Interview & 83 & 117 & 104 & 13  \\
YouTube Vlog & 191 & 337 & 318 & 19  \\
\midrule
Total & 630 & 1,000 & 890 & 110  \\
\bottomrule
\end{tabular}
\end{table*}

\begin{table}[!htbp]
\centering
\caption{Composition of the PsycheMME benchmark: 1,000 questions from 630 episodes. G/W/N/T/S/A/M/R denote the capabilities defined in Appendix~\ref{app:taxonomy_details}; Choice and Open denote question formats.}
\label{tab:app_denominators}
\begingroup
\small
\setlength{\tabcolsep}{4pt}
\renewcommand{\arraystretch}{1.12}
\psycheapptablefit{%
\begin{tabular}{lrrr}
\toprule
Cap. & $n$ & Choice & Open \\
\midrule
G & 186 & 184 & 2  \\
W & 320 & 287 & 33  \\
N & 97 & 62 & 35  \\
T & 72 & 72 & 0  \\
S & 97 & 64 & 33  \\
A & 94 & 87 & 7  \\
M & 61 & 61 & 0  \\
R & 73 & 73 & 0  \\
\midrule
Total & 1000 & 890 & 110  \\
\bottomrule
\end{tabular}
}
\endgroup
\end{table}

The video sources include Ego4D~\cite{grauman2022ego4d},
EgoCom~\cite{northcutt2020egocom}, MERCI+~\cite{althubyani2025merci},
CASTLE~\cite{rossetto2025castle}, EasyCom~\cite{donley2021easycom},
and IndEgo~\cite{chavan2025indego}.

Episode selection prioritizes social relevance, observable behavioral
cues, and adequate video quality. Each question identifies a target
person, interaction, and relevant time window, with its reference answer
grounded in the source cues. For example, speaker labels require
visual corroboration before they can establish a person's identity;
psychological causes and preferences likewise require contextual support.
These criteria implement the requirements for cue grounding in
Appendix~\ref{app:taxonomy_evidence}.

\subsection{Question Construction with Independent Answering}
\label{app:construction_details}

Question construction separates proposing a reference answer from
independently answering the question
(Fig.~\ref{fig:app_construction_workflow}). The Question Agent proposes
a question, candidate answer, applicable options, and supporting cues.
The Answer Agent then examines the same video using only the question,
options, and public context. With the proposed answer and its supporting
rationale withheld, this agent provides an independent answer or
identifies ambiguity and insufficient observable support.

\psycheappfigure{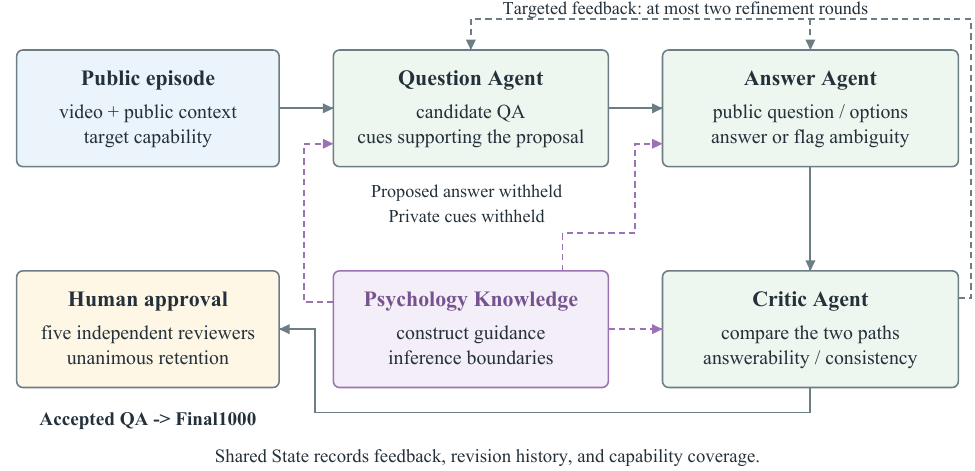}{fig:app_construction_workflow}{Question construction and review. The Answer Agent independently examines the episode without access to the proposed answer or its supporting cues. Psychology Knowledge guides the three agents' judgments, and critic feedback supports up to two rounds of refinement. Final inclusion requires unanimous human approval.}

\begin{table*}[!t]
\centering
\caption{Roles of the three construction agents. Psychology Knowledge guides their interpretation of the target constructs. The Answer Agent receives only the video, question, options, and public context.}
\label{tab:app_agent_access}
\small
\setlength{\tabcolsep}{4pt}
\renewcommand{\arraystretch}{1.12}
\begin{tabular}{@{}p{.18\textwidth}p{.30\textwidth}p{.24\textwidth}p{.22\textwidth}@{}}
\toprule
Agent & Inputs & Assessment & Output \\
\midrule
Question Agent & Video, public context, target capability & Observed cues support the question and reference & Candidate question, answer, options, and supporting cues \\
Answer Agent & Video, question, options, public context & Answerability without the proposed reference or its supporting cues & Independent answer or identified ambiguity \\
Critic Agent & Proposed and independent answers, source cues & Cue support, logical coherence, construct alignment & Quality assessment and refinement suggestions \\
\bottomrule
\end{tabular}
\end{table*}

The Critic Agent evaluates answerability, cue support, and logical
coherence by comparing the proposed and independent answers,
following the use of multiple answers in multiagent
reasoning~\cite{du2024debate}. Psychology
Knowledge guides all three agents in aligning questions with the target
construct and distinguishing observable cues from inferred states,
needs, and actions. Candidate acceptance combines cue support, logical
coherence, alignment with the target capability, and agreement between
answers.

\subsection{Iterative Question Refinement}
\label{app:refinement_details}

Critic feedback determines whether a candidate needs a clearer question
or a renewed examination of the cues. Ambiguous referents,
unsupported premises, and defects in answer options prompt question revision;
reasoning errors prompt another independent answer. Every revised
question is answered again without revealing the proposed reference.
Candidates that remain unresolved after two refinement rounds are
discarded.

Shared State coordinates feedback and tracks capability coverage during
construction. Coverage guides further question generation toward
underrepresented capabilities, while the same answerability and
criteria for cue support apply to every candidate.

\subsection{Human Quality Review}
\label{app:quality_details}

As described in Section~\ref{sec:benchmark_construction}, all benchmark
items require unanimous approval from five researchers with psychology
backgrounds. Reviewers check question validity, reference correctness,
and psychological consistency using the capability definitions and
cue requirements in Table~\ref{tab:app_review_codebook}.

\begin{table*}[!t]
\centering
\caption{Criteria for expert question review by capability. Each target judgment requires support from the available episode and public context.}
\label{tab:app_review_codebook}
\small
\setlength{\tabcolsep}{5pt}
\renewcommand{\arraystretch}{1.12}
\begin{tabular}{@{}p{.10\textwidth}p{.42\textwidth}p{.43\textwidth}@{}}
\toprule
Capability & Required cues & Review criterion \\
\midrule
G & A uniquely grounded speaker, addressee, person, or interpersonal relation & Corroborate the target person's visible identity using speech, behavior, and interaction relations. \\
W & Observable cues supporting a psychological state or its stated cause & Distinguish situated psychological interpretation from clinical diagnosis; check the support for each stated cause. \\
N & An explicit request, refusal, or contextual basis for the inferred support need & Check whether the available cues indicate willingness to receive the proposed support. \\
T & Turn structure, ongoing activity, urgency, or a supported opportunity to respond & Assess the intervention opportunity against turn structure, current activity, and urgency. \\
S & Response content and tone suitable for the supported state, need, and context & Assess the fit of content, strategy, and tone to the person's state, need, and context. \\
A & A feasible physical or nonverbal response consistent with the interaction & Distinguish recognition of observed behavior from selection of an appropriate next action. \\
M & Relevant history available in the supplied context & Ground personal preferences and prior interactions in the supplied history. \\
R & Cues concerning consent, refusal, boundaries, autonomy, or possible harm & Check consent, expressed refusals, personal boundaries, and consequences before endorsing intervention. \\
\bottomrule
\end{tabular}
\end{table*}

\section{Experimental and Evaluation Protocol}
\label{app:experimental_details}

\subsection{Models and Analysis Cohorts}
\label{app:model_details}

The foundation comparison evaluates 17 models: eight proprietary models and nine open-weight models.
The open-weight group contains Qwen3.8-27B, Qwen3.5-9B,
Qwen3.5-4B Base, InternVL3.5-14B/8B/4B, Gemma-4-12B-it,
Phi-4-multimodal, and MiniCPM-V4.5. The proprietary group comprises
Claude Opus 5, Gemini 3.1 Pro, Gemini 3.6 Flash, GLM-5V-Turbo,
GPT-5.6 Sol, Kimi K3, Qwen3.8-Max, and Seed 2.1 Pro.
Table~\ref{tab:foundation_main} reports these models and two adapted checkpoints.
Qwen3.5-4B's SFT and GRPO variants are examined separately in
Appendix~\ref{app:post_training_details}.

\subsection{Model Inputs and Response Format}
\label{app:input_details}

The evaluation set contains 1,000 questions from 630 episodes:
890 choice questions and 110 open questions.
The G/W/N/T/S/A/M/R counts are 186/320/97/72/97/94/61/73.
We prioritize native video and native audio independently, according
to the evaluated model interface. Sampled frames are the fallback
for video, using 1 FPS with at most 64 frames; available ASR transcripts
are the fallback for audio. Native inputs are used wherever the
corresponding input path is supported and available.

Gemini receives native video and audio; Claude and GPT use sampled frames with ASR transcripts; GLM, Qwen-Max, and Seed receive native video with ASR. Kimi primarily uses native video with ASR, supplemented by sampled frames when needed. The three diagnostic open-weight models are evaluated using ASR transcripts where available.

\paragraph{Common input policy and paired controls.}
Input formats follow the media interfaces supported by each model.
The policy preserves native video and audio where available and
standardizes the fallbacks through a common frame sampling rule and
the available ASR transcripts. All models answer the same benchmark
questions under a shared response schema and scoring rubric, using
the same reference answers. Table~\ref{tab:foundation_main} reports
performance under this protocol.
The paired controls in Appendix~\ref{app:modality_analysis} examine
modality availability and visual coverage on the
same questions; Appendix~\ref{app:memory_analysis} examines relevant
history while holding the video, question, and remaining inputs fixed.

Models provide an answer and eight rationale fields, G, W, N, T, S,
A, M, and R. Unused fields may be null; rationale coverage requires
at least one informative explanation rather than completion of every
field. Qwen3.5-9B uses temperature 0, a 768-token output limit,
disabled thinking, and responses formatted as JSON. InternVL3.5-8B uses
temperature 0, the same output limit, and one image tile per frame.
Qwen3.5-4B Base uses greedy BF16 inference under the input and scoring
protocol in Appendix~\ref{app:post_training_details}.
The Judge is Qwen3.8-27B.

\subsection{Evaluation Instructions and Response Format}
\label{app:response_contract}

Models answer questions using native video and native audio where
supported, with sampled frames and ASR transcripts as the respective
fallback inputs (Appendix~\ref{app:input_details}). They also receive
answer options and relevant public context.
Reference answers and private cue annotations are provided only
to the evaluator after inference
(Fig.~\ref{fig:app_scoring_workflow}).

\psycheappfigure{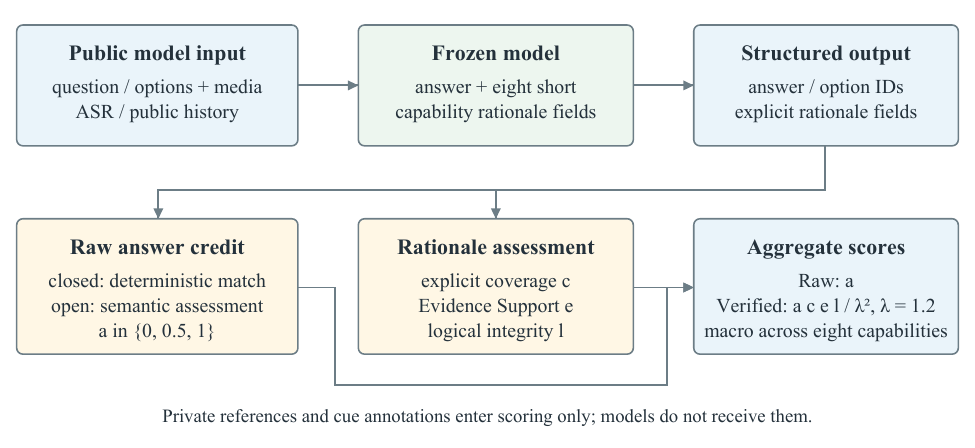}{fig:app_scoring_workflow}{Evaluation and scoring procedure. Reference answers and cue annotations are provided only to the evaluator. Answer correctness and rationale quality are assessed separately before computing Verified credit.}

The response format uses one JSON object with the keys
\texttt{answer}, \texttt{G}, \texttt{W}, \texttt{N}, \texttt{T},
\texttt{S}, \texttt{A}, \texttt{M}, and \texttt{R}.
Choice answers use an option ID when one option is required, or an
array of option IDs when several are required. Open answers use
concise natural language. Choice answers must match the reference
option or complete option set exactly: missing a correct option or
selecting an extra option receives zero answer credit.

Each capability field contains a short judgment contributing to the
answer, or JSON \texttt{null} when unused. At least one informative
rationale is required; empty and null fields do not provide
rationale coverage.
\begin{table}[!htbp]
\centering
\caption{Structured response fields and answer representations. The eight rationale fields are short explicit judgments; unused fields are JSON null.}
\label{tab:app_response_schema}
\small
\setlength{\tabcolsep}{4pt}
\renewcommand{\arraystretch}{1.12}
\begin{tabular}{@{}p{.34\linewidth}p{.61\linewidth}@{}}
\toprule
Field / task & Required representation \\
\midrule
\texttt{answer}: choice & Option ID for one required option, e.g., \texttt{"A"}; array for several, e.g., \texttt{["A", "C"]} \\
\texttt{answer}: open & Concise answer in natural language \\
\texttt{G/W/N/T/S/A/M/R} & Short relevant judgment or JSON \texttt{null} \\
\bottomrule
\end{tabular}
\end{table}

The following example illustrates a choice response requiring one option.
When several options are required, the answer is an array such as
\texttt{["A", "C"]}; for open questions, it is concise natural language.
\begin{quote}\small
\begin{minipage}{\linewidth}
\begin{verbatim}
{"answer": "A",
 "G": "Brief grounded observation.",
 "W": null, "N": null, "T": null,
 "S": null, "A": null,
 "M": null, "R": null}
\end{verbatim}
\end{minipage}
\end{quote}

\subsection{Judge Instructions and Rubric}
\label{app:judge_rubric}

The Judge receives the question, applicable options,
reference answer, textual descriptions of reference cues, and the
model's answer and rationales. It assesses explanation support
against these cues and checks logical consistency.
Reasoning-Verified measures the quality of explicit justifications
relative to annotated reference cues and the final answer.
Multiple psychological interpretations are accepted when supported
by the available cues.

Answer correctness and rationale quality are assessed separately.
Open answers receive semantic credit in $\{0,0.5,1\}$:
full credit for a correct answer, partial credit for a materially
correct but incomplete answer, and zero for incorrect or unsupported
answers or responses that do not address the question.

Evidence Support concerns whether the explanation's observations and
inferences are grounded in the reference cue annotations. Logical Integrity
concerns whether the active rationale fields are mutually compatible
and support the final answer. An internally consistent explanation
may lack support from observed cues, while grounded statements may
combine into an inconsistent conclusion. Rationale coverage is evaluated separately.
The Judge assigns Evidence Support and Logical Integrity labels
with a short explanation for each assessment.
\begin{table*}[!t]
\centering
\caption{Interpretation of rationale labels and their scoring multipliers. Evidence Support and Logical Integrity are separate judgments.}
\label{tab:app_judge_rubric}
\small
\setlength{\tabcolsep}{5pt}
\renewcommand{\arraystretch}{1.13}
\begin{tabular}{@{}p{.17\textwidth}p{.35\textwidth}p{.35\textwidth}r@{}}
\toprule
Support / logic label & Interpretation of Evidence Support & Interpretation of Logical Integrity & Weight \\
\midrule
Strong / fully coherent & Relevant claims are well supported by frozen cue annotations. & Active judgments jointly support the answer without contradiction. & 1.2 \\
Normal / acceptable & Support is adequate, with limited omissions or specificity. & The explanation is consistent and adequate, with minor gaps. & 1.0 \\
Weak / weak & A material inference or supporting link lacks adequate support from observed cues. & A material reasoning link is incomplete or inadequately justified. & 0.8 \\
Contradicted / contradictory & A material claim conflicts with the supplied cues. & Active judgments contradict one another or the final answer. & 0.5 \\
Insufficient / unevaluable & No evaluable support is available for the explicit rationale. & No evaluable logical explanation is present. & 0.0 \\
\bottomrule
\end{tabular}
\end{table*}

Table~\ref{tab:app_scoring_examples} illustrates how answer credit,
rationale coverage, and the two quality labels determine Verified credit.
\begin{table}[!htbp]
\centering
\caption{Arithmetic examples of Reasoning-Verified credit for individual questions. Each row specifies $(a,c,e,\ell)$; the final column is $100ace\ell/\lambda^2$, with $\lambda=1.2$.}
\label{tab:app_scoring_examples}
\small
\setlength{\tabcolsep}{5pt}
\renewcommand{\arraystretch}{1.10}
\begin{tabular}{lrrrrr}
\toprule
Case & $a$ & $c$ & $e$ & $\ell$ & Verified (\%) \\
\midrule
Full support & 1 & 1 & 1.2 & 1.2 & 100.00 \\
Normal / acceptable & 1 & 1 & 1.0 & 1.0 & 69.44 \\
Weak support & 1 & 1 & 0.8 & 1.2 & 66.67 \\
Contradictory logic & 1 & 1 & 1.2 & 0.5 & 41.67 \\
Missing rationale & 1 & 0 & 0 & 0 & 0.00 \\
Incorrect answer & 0 & 1 & 1.2 & 1.2 & 0.00 \\
\bottomrule
\end{tabular}
\end{table}

\subsection{Scoring and Statistical Estimation}
\label{app:scoring_details}
\label{app:evaluation_details}

Raw credit $a_q$ evaluates choice answers through exact matching of
the selected option or complete option set, and assigns semantic
credit in $\{0,0.5,1\}$ for open questions.
Malformed or unparseable answers and responses without an answer
receive zero credit and remain in the evaluation denominator.
Verified credit combines answer correctness with rationale quality:
\begin{equation}
v_q=a_qc_qe_q\ell_q/\lambda^2,\quad \lambda=1.2,
\label{eq:app_verified_score}
\end{equation}
where $c_q$ indicates whether an informative rationale is present,
and $e_q$ and $\ell_q$ represent Evidence Support and Logical Integrity.
The gap between Raw and Verified scores measures the reduction in answer credit under
the rationale rubric.

\paragraph{Rubric Weights and Normalization.}
Evidence Support labels strong/normal/weak/contradicted receive
1.2/1.0/0.8/0.5; fully coherent/acceptable/weak/contradictory
logic labels receive the same multipliers
(Table~\ref{tab:app_judge_rubric}). Insufficient evidence,
unevaluable logic, and absent rationales
receive zero Verified credit. The evaluable labels have weights in
$[0.5,\lambda]$, with $\lambda=1.2$ as the maximum in each dimension;
including unevaluable labels gives $e_q,\ell_q\in[0,\lambda]$.

The multipliers center on 1.0 for normal cue support and acceptable
logic, with symmetric steps of 0.2 distinguishing strong (1.2) and weak
(0.8) judgments and a steeper discount (0.5) for contradictions.
Dividing $e_q\ell_q$
by $\lambda^2$ yields 1 when both dimensions receive their
strongest labels. Since $a_q,c_q\in[0,1]$, this also ensures
$0\leq v_q\leq a_q\leq1$.
A correct answer with an informative rationale, normal Evidence Support,
and acceptable Logical Integrity earns $1/1.44\approx69.44\%$ credit.
Alternative Soft and Strict weights preserve the ordering of the
three diagnostic models (Appendix~\ref{app:weight_sensitivity}).

\paragraph{Aggregation and Credit Retention.}
Macro scores on the full benchmark give equal weight to the eight capabilities
$\mathcal C$, for either Raw ($x=a$) or Verified ($x=v$) credit:
\begin{equation}
\operatorname{Macro}(x_m)=\frac{100}{8}
\sum_{k\in\mathcal C}\frac{1}{n_k}\sum_{q\in k}x_{mq}.
\end{equation}
Micro scores instead average questions within the named subset.
For model $m$ and capability $k$, credit retention is
\begin{equation}
\rho_{mk}=\frac{\sum_{q\in k}v_{mq}}{\sum_{q\in k}a_{mq}}.
\label{eq:app_retention}
\end{equation}
It measures the proportion of earned answer credit retained after
verification. For binary answer credit, it equals the mean verification
weight among correct answers. Pooled retention similarly divides
total Verified credit by total Raw credit. Conditional Evidence Support and
Logical Integrity frequencies characterize the reasons for reduced credit.

\paragraph{Paired Episode Bootstrap.}
Confidence intervals resample episodes while retaining their questions
and pairing across models or conditions. We use 10,000 draws with
seed 42 and report percentile 95\% confidence intervals.
Paired differences are recomputed within each draw; undefined draws
are excluded where applicable. Intervals capture episode sampling
variation; post hoc comparisons are exploratory and unadjusted unless
otherwise specified.

\paragraph{Stage Credit and Loss Attribution.}
The successive stages in Fig.~\ref{fig:task_progress}(b,d) correspond to
$a$, $ac$, $ac\tilde e$, and $v=ac\tilde e\tilde\ell$,
where $\tilde e=e/\lambda$ and $\tilde\ell=\ell/\lambda$.
Their successive losses are $1-a$, $a(1-c)$,
$ac(1-\tilde e)$, and $ac\tilde e(1-\tilde\ell)$.
Together with $v$, they sum to one for each response.
Evidence Support precedes Logical Integrity in this sequential credit decomposition,
making their allocated losses dependent on their order. Stage credit uses
capability macro weights; distribution bins weight questions equally.

\section{SFT and GRPO on PsycheMME}
\label{app:post_training_details}
\label{app:post_training}
\label{app:adaptation_analysis}

\subsection{SFT and GRPO Training Configurations}
\label{app:training_recipe}

Both adaptation stages update LoRA parameters in the language backbone of
Qwen3.5-4B while freezing the visual encoder and multimodal aligner.
LoRA is applied to the linear projections in the language model with rank 8,
scaling $\alpha=32$, and dropout 0.05.

SFT trains on 3,657 video QA examples for one epoch (458 updates).
Each example contains 32 sampled frames with a maximum image dimension
of 256 pixels, ASR, and public history where available. Targets contain the
answer and one annotated observation; unused rationale fields are null.
The loss is computed over assistant tokens. Training uses BF16,
fused AdamW, a constant learning rate of $5\times10^{-6}$,
weight decay 0.1, no warmup, a batch size of one with eight gradient
accumulation steps, and a maximum sequence length of 4,096 tokens.

GRPO starts from the SFT checkpoint and uses it as a fixed reference.
Its training pool contains 2,842 single-choice examples with the same
32-frame representation. For each question, eight responses are sampled
at temperature 1.0 and top-$p=0.9$, with at most 768 completion tokens.
The binary reward in Eq.~\ref{eq:grpo_reward} requires both exact
answer correctness and compliance with the response requirements. Evidence Support and Logical Integrity
scores are used only for evaluation.

The GRPO objective uses rewards normalized within each group, clipping
$\epsilon=0.2$, and KL regularization with $\beta=0.03$ against the
fixed SFT reference. Training uses a constant learning rate of
$5\times10^{-7}$, no warmup, and seed 42. Groups with constant rewards
or truncated responses are excluded. Each update uses eight distinct
questions and 64 sampled responses, giving 32 updates over
256 distinct questions.

\subsection{Checkpoints and Evaluation Scope}
\label{app:post_training_scope}

The adaptation comparison evaluates Qwen3.5-4B Base, SFT at step 458,
and GRPO at step 32 on the same 1,000 benchmark questions.
Evaluation uses 1 FPS with at most 64 frames, without the reduction in image size used during training, together with greedy BF16 inference
and a 768-token output limit. All three models use the same
rules for scoring answers and Qwen3.8-27B rationale Judge.

Raw/Verified macro scores are 47.68/37.71 for Base, 72.58/52.13
for SFT, and 73.19/52.62 for GRPO
(Table~\ref{tab:psychebench-sft-grpo}).
Figure~\ref{fig:app_training_comparison} shows the overall scores and
Raw profiles across the eight capabilities.
SFT improves several capabilities, but Timing remains below the Base
score after both adaptation stages. This comparison does not isolate
the effects of temporal sampling from the training objectives.

\psycheappfigure{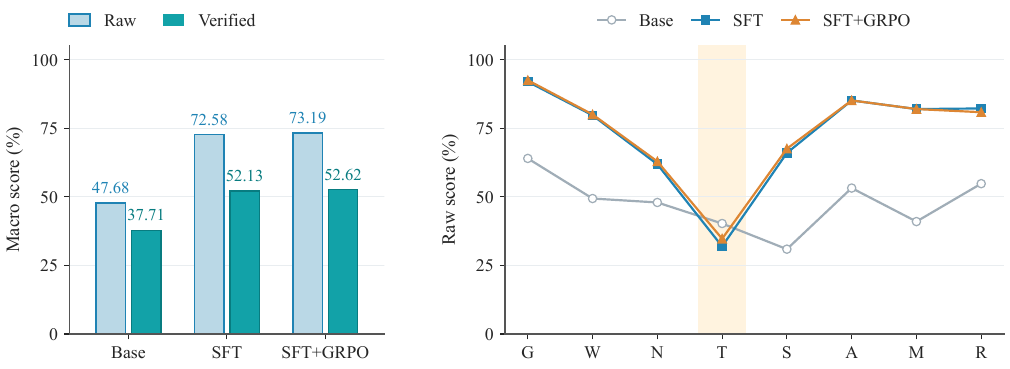}{fig:app_training_comparison}{Qwen3.5-4B adaptation on the same 1,000 PsycheMME questions with greedy decoding. Left: overall Raw and Verified macro scores for Base, SFT at step 458, and SFT+GRPO at step 32. Right: Raw scores across the eight capabilities, with Response Timing shaded. Capability abbreviations follow Table~\ref{tab:psychebench-sft-grpo}.}

\subsection{Training and Evaluation Data}
\label{app:post_training_limits}

To prevent data leakage, we exclude benchmark questions and video episodes from training. We further split the data by source video and remove clips that share the same source with the development and held-out sets. This ensures that the SFT training data does not overlap with the benchmark or checkpoint screening set.

Checkpoint screening and final evaluation use independent
question sets: the screening set is used for model selection,
and the evaluation set is used to report the results in
Table~\ref{tab:psychebench-sft-grpo}. This adaptation comparison is
separate from the 17-model foundation evaluation.

\begin{table*}[t]
\centering
\caption{Base, SFT, and GRPO results on PsycheMME
(\%, 1,000 questions per model).
Overall Raw and Ver.\ denote raw accuracy and Verified scores,
respectively, averaged equally over the eight capabilities.
Capability columns report Raw scores.
Blue shading indicates score magnitude;
bold denotes the highest observed value in each column,
including ties.}
\label{tab:psychebench-sft-grpo}
\begingroup
\small
\setlength{\tabcolsep}{5pt}
\renewcommand{\arraystretch}{1.15}

\definecolor{PBResultBlue}{RGB}{181,211,232}
\newcommand{\pbscore}[1]{%
  \cellcolor{PBResultBlue!#1}#1%
}
\newcommand{\pbbest}[1]{%
  \cellcolor{PBResultBlue!#1}\textbf{#1}%
}

\resizebox{\textwidth}{!}{%
\begin{tabular}{@{}l*{10}{c}@{}}
\toprule
\textbf{Model}
& \multicolumn{2}{c}{\textbf{Overall}}
& \multicolumn{3}{c}{\textbf{Understanding}}
& \multicolumn{3}{c}{\textbf{Interaction}}
& \multicolumn{2}{c}{\textbf{Contextual}} \\
\cmidrule(lr){2-3}
\cmidrule(lr){4-6}
\cmidrule(lr){7-9}
\cmidrule(lr){10-11}
& Raw & Ver.
& G & W
& N & T & S & A
& M & R \\
\midrule

Base (Qwen3.5-4B)
& \pbscore{47.68} & \pbscore{37.71}
& \pbscore{63.98} & \pbscore{49.38}
& \pbscore{47.94} & \pbbest{40.28}
& \pbscore{30.93} & \pbscore{53.19}
& \pbscore{40.98} & \pbscore{54.79} \\

SFT (step 458)
& \pbscore{72.58} & \pbscore{52.13}
& \pbscore{91.94} & \pbscore{79.69}
& \pbscore{61.86} & \pbscore{31.94}
& \pbscore{65.98} & \pbbest{85.11}
& \pbbest{81.97} & \pbbest{82.19} \\

GRPO (step 32)
& \pbbest{73.19} & \pbbest{52.62}
& \pbbest{92.47} & \pbbest{80.00}
& \pbbest{62.89} & \pbscore{34.72}
& \pbbest{67.53} & \pbbest{85.11}
& \pbbest{81.97} & \pbscore{80.82} \\

\bottomrule
\end{tabular}%
}
\endgroup

\end{table*}

\section{Complete Foundation Model Results}
\label{app:foundation_statistics}

\subsection{Capability Profiles}
\label{app:capability_profiles}
Tables~\ref{tab:app_raw} and~\ref{tab:app_verified} report
capability scores for all 17 foundation models with 95\% confidence
intervals from bootstrapping episodes.
All eight proprietary models exceed 88\% Verified in Grounding,
while Need, Timing, and Action show wider differences across models.
Gemini 3.6 exceeds Gemini Pro by 5.70 Verified macro points
(95\% CI $[3.57,7.84]$), but the difference is uneven:
0.46 in G, 9.67 in N, 11.10 in S, and 12.78 in A.
The relative advantage in A versus G is 12.32 points
($[5.40,19.67]$), indicating different capability profiles rather
than a uniform improvement.

Kimi and GPT have nearly identical Verified macro scores,
75.35 and 75.98 (paired difference -0.63; $[-2.97,1.67]$).
Differences between Kimi and GPT are $+4.52$ in S and $-6.66$ in N;
their relative difference in S versus N is 11.18
($[1.26,21.41]$). The individual S and N intervals include zero.
GPT exceeds Gemini Pro in overall point estimate but is lower
in T by 11.73 points ($[-21.24,-2.23]$).

Figure~\ref{fig:app_capability_heatmap} visualizes the Verified profiles
for 16 foundation models; Qwen3.5-4B is shown separately in
Figure~\ref{fig:app_training_comparison}.

\psycheappfigure{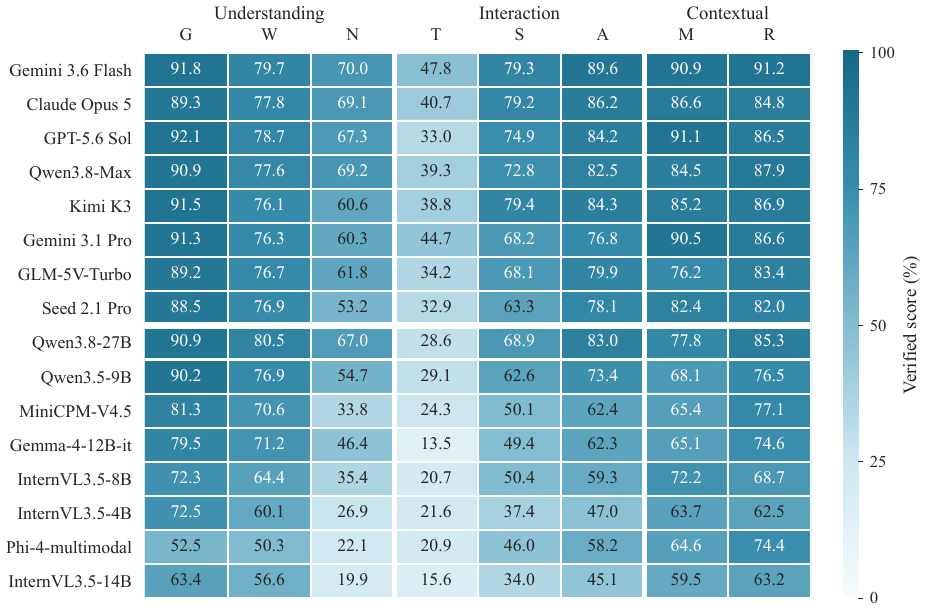}{fig:app_capability_heatmap}{Verified scores (\%) for 16 foundation models on 1,000 PsycheMME questions, sorted by Verified macro score within each model group. G: Grounding; W: Psychological Understanding; N: Need; T: Response Timing; S: Verbal Response; A: Action Decision; M: Memory; R: Boundary and Risk. All cells share a 0 to 100 color scale; the gold outline marks Response Timing.}

\begin{table*}[tp]
\centering
\caption{Raw scores by capability and model (\%), with 95\% confidence intervals from the episode bootstrap.}
\label{tab:app_raw}
\small
\setlength{\tabcolsep}{3.2pt}
\renewcommand{\arraystretch}{0.99}
\psycheapptablefit{%
\begin{tabular}{lrrrrrrrr}
\toprule
Model & G & W & N & T & S & A & M & R \\
\midrule
Claude Opus 5 & \shortstack{91.94\\{\scriptsize [87.89,95.65]}} & \shortstack{79.84\\{\scriptsize [75.54,83.93]}} & \shortstack{76.80\\{\scriptsize [68.75,84.27]}} & \shortstack{54.17\\{\scriptsize [42.42,65.67]}} & \shortstack{82.47\\{\scriptsize [74.48,89.64]}} & \shortstack{89.36\\{\scriptsize [83.12,94.91]}} & \shortstack{95.08\\{\scriptsize [89.09,100.00]}} & \shortstack{89.04\\{\scriptsize [81.69,95.65]}} \\
Gemini 3.1 Pro & \shortstack{93.55\\{\scriptsize [90.00,96.80]}} & \shortstack{79.22\\{\scriptsize [74.93,83.43]}} & \shortstack{67.53\\{\scriptsize [58.85,76.04]}} & \shortstack{66.67\\{\scriptsize [55.26,77.27]}} & \shortstack{71.65\\{\scriptsize [62.73,79.91]}} & \shortstack{83.51\\{\scriptsize [75.82,90.53]}} & \shortstack{98.36\\{\scriptsize [94.44,100.00]}} & \shortstack{90.41\\{\scriptsize [83.10,96.77]}} \\
Gemini 3.6 Flash & \shortstack{93.82\\{\scriptsize [90.24,96.94]}} & \shortstack{82.19\\{\scriptsize [78.08,86.08]}} & \shortstack{76.29\\{\scriptsize [69.07,83.33]}} & \shortstack{66.67\\{\scriptsize [55.71,77.22]}} & \shortstack{82.99\\{\scriptsize [76.09,89.24]}} & \shortstack{93.62\\{\scriptsize [88.64,97.89]}} & \shortstack{96.72\\{\scriptsize [91.30,100.00]}} & \shortstack{93.15\\{\scriptsize [86.75,98.53]}} \\
GLM-5V-Turbo & \shortstack{91.67\\{\scriptsize [87.50,95.41]}} & \shortstack{79.38\\{\scriptsize [75.16,83.39]}} & \shortstack{71.13\\{\scriptsize [62.78,79.25]}} & \shortstack{50.00\\{\scriptsize [38.46,61.73]}} & \shortstack{73.20\\{\scriptsize [64.36,81.25]}} & \shortstack{87.23\\{\scriptsize [80.21,93.55]}} & \shortstack{86.89\\{\scriptsize [77.05,95.38]}} & \shortstack{87.67\\{\scriptsize [79.73,94.67]}} \\
GPT-5.6 Sol & \shortstack{93.82\\{\scriptsize [90.22,97.01]}} & \shortstack{80.62\\{\scriptsize [76.45,84.63]}} & \shortstack{72.16\\{\scriptsize [63.83,80.21]}} & \shortstack{48.61\\{\scriptsize [36.99,60.27]}} & \shortstack{78.87\\{\scriptsize [70.89,85.94]}} & \shortstack{87.23\\{\scriptsize [80.00,93.48]}} & \shortstack{98.36\\{\scriptsize [94.44,100.00]}} & \shortstack{89.04\\{\scriptsize [81.48,95.59]}} \\
Kimi K3 & \shortstack{92.74\\{\scriptsize [88.83,96.26]}} & \shortstack{77.50\\{\scriptsize [73.05,81.75]}} & \shortstack{68.04\\{\scriptsize [59.18,76.58]}} & \shortstack{51.39\\{\scriptsize [39.73,62.65]}} & \shortstack{83.51\\{\scriptsize [76.11,90.10]}} & \shortstack{87.77\\{\scriptsize [81.05,93.89]}} & \shortstack{93.44\\{\scriptsize [86.67,98.57]}} & \shortstack{90.41\\{\scriptsize [83.13,96.67]}} \\
Qwen3.8-Max & \shortstack{91.94\\{\scriptsize [87.88,95.45]}} & \shortstack{79.22\\{\scriptsize [74.84,83.33]}} & \shortstack{76.29\\{\scriptsize [68.50,83.64]}} & \shortstack{52.78\\{\scriptsize [40.85,64.41]}} & \shortstack{77.84\\{\scriptsize [69.19,85.65]}} & \shortstack{88.30\\{\scriptsize [81.67,94.44]}} & \shortstack{95.08\\{\scriptsize [89.16,100.00]}} & \shortstack{90.41\\{\scriptsize [83.13,96.67]}} \\
Seed 2.1 Pro & \shortstack{91.67\\{\scriptsize [87.57,95.48]}} & \shortstack{80.78\\{\scriptsize [76.70,84.80]}} & \shortstack{59.28\\{\scriptsize [50.00,68.24]}} & \shortstack{48.61\\{\scriptsize [36.92,60.29]}} & \shortstack{69.07\\{\scriptsize [60.00,77.62]}} & \shortstack{85.11\\{\scriptsize [77.72,91.95]}} & \shortstack{90.16\\{\scriptsize [80.95,97.92]}} & \shortstack{86.30\\{\scriptsize [78.08,93.75]}} \\
Qwen3.8-27B & \shortstack{93.55\\{\scriptsize [90.00,96.74]}} & \shortstack{82.34\\{\scriptsize [78.39,86.17]}} & \shortstack{75.26\\{\scriptsize [67.16,83.01]}} & \shortstack{45.83\\{\scriptsize [34.25,57.14]}} & \shortstack{74.74\\{\scriptsize [66.82,82.18]}} & \shortstack{87.77\\{\scriptsize [81.01,94.02]}} & \shortstack{88.52\\{\scriptsize [78.18,96.77]}} & \shortstack{91.78\\{\scriptsize [85.06,97.33]}} \\
Qwen3.5-9B & \shortstack{93.55\\{\scriptsize [89.95,96.80]}} & \shortstack{79.84\\{\scriptsize [75.47,84.01]}} & \shortstack{69.59\\{\scriptsize [60.78,78.19]}} & \shortstack{48.61\\{\scriptsize [36.59,60.27]}} & \shortstack{72.16\\{\scriptsize [63.59,80.00]}} & \shortstack{80.32\\{\scriptsize [72.15,88.04]}} & \shortstack{81.97\\{\scriptsize [70.59,91.67]}} & \shortstack{86.30\\{\scriptsize [78.05,93.65]}} \\
InternVL3.5-14B & \shortstack{68.01\\{\scriptsize [61.29,74.61]}} & \shortstack{62.50\\{\scriptsize [57.28,67.61]}} & \shortstack{29.38\\{\scriptsize [21.23,37.89]}} & \shortstack{36.11\\{\scriptsize [25.32,47.56]}} & \shortstack{36.60\\{\scriptsize [27.00,46.15]}} & \shortstack{54.26\\{\scriptsize [44.59,64.09]}} & \shortstack{63.93\\{\scriptsize [51.22,75.51]}} & \shortstack{67.12\\{\scriptsize [56.25,77.63]}} \\
Gemma-4-12B-it & \shortstack{84.95\\{\scriptsize [79.65,89.89]}} & \shortstack{75.94\\{\scriptsize [71.31,80.28]}} & \shortstack{57.73\\{\scriptsize [48.45,66.85]}} & \shortstack{25.00\\{\scriptsize [15.28,35.14]}} & \shortstack{57.73\\{\scriptsize [48.02,67.00]}} & \shortstack{72.34\\{\scriptsize [63.07,81.13]}} & \shortstack{75.41\\{\scriptsize [63.64,86.30]}} & \shortstack{82.19\\{\scriptsize [73.13,90.41]}} \\
InternVL3.5-8B & \shortstack{79.84\\{\scriptsize [74.01,85.43]}} & \shortstack{70.47\\{\scriptsize [65.54,75.31]}} & \shortstack{52.58\\{\scriptsize [43.96,61.00]}} & \shortstack{33.33\\{\scriptsize [22.62,44.44]}} & \shortstack{61.86\\{\scriptsize [52.36,70.62]}} & \shortstack{70.74\\{\scriptsize [61.58,79.61]}} & \shortstack{86.89\\{\scriptsize [78.05,94.55]}} & \shortstack{83.56\\{\scriptsize [74.67,91.80]}} \\
Phi-4-multimodal & \shortstack{66.13\\{\scriptsize [59.22,72.78]}} & \shortstack{60.47\\{\scriptsize [55.38,65.65]}} & \shortstack{31.96\\{\scriptsize [23.40,40.56]}} & \shortstack{43.06\\{\scriptsize [31.43,54.43]}} & \shortstack{61.86\\{\scriptsize [52.25,71.18]}} & \shortstack{71.28\\{\scriptsize [62.02,80.37]}} & \shortstack{75.41\\{\scriptsize [64.06,86.27]}} & \shortstack{78.08\\{\scriptsize [68.42,87.30]}} \\
InternVL3.5-4B & \shortstack{84.14\\{\scriptsize [78.89,89.27]}} & \shortstack{70.16\\{\scriptsize [65.23,75.00]}} & \shortstack{42.78\\{\scriptsize [34.02,51.58]}} & \shortstack{36.11\\{\scriptsize [25.00,47.22]}} & \shortstack{45.88\\{\scriptsize [36.63,55.00]}} & \shortstack{62.77\\{\scriptsize [52.83,72.53]}} & \shortstack{70.49\\{\scriptsize [59.62,81.08]}} & \shortstack{78.08\\{\scriptsize [68.35,87.14]}} \\
MiniCPM-V4.5 & \shortstack{87.37\\{\scriptsize [82.35,91.85]}} & \shortstack{78.12\\{\scriptsize [73.77,82.32]}} & \shortstack{47.42\\{\scriptsize [38.46,56.38]}} & \shortstack{43.06\\{\scriptsize [31.67,54.55]}} & \shortstack{57.73\\{\scriptsize [48.26,66.86]}} & \shortstack{76.60\\{\scriptsize [67.89,84.66]}} & \shortstack{81.97\\{\scriptsize [69.70,92.45]}} & \shortstack{84.93\\{\scriptsize [76.39,92.73]}} \\
Qwen3.5-4B & \shortstack{63.98\\{\scriptsize [57.07,70.67]}} & \shortstack{49.38\\{\scriptsize [44.01,54.90]}} & \shortstack{47.94\\{\scriptsize [38.83,57.22]}} & \shortstack{40.28\\{\scriptsize [29.03,51.69]}} & \shortstack{30.93\\{\scriptsize [22.22,39.71]}} & \shortstack{53.19\\{\scriptsize [43.23,63.25]}} & \shortstack{40.98\\{\scriptsize [26.32,56.86]}} & \shortstack{54.79\\{\scriptsize [43.08,66.67]}} \\
\bottomrule
\end{tabular}}
\end{table*}

\begin{table*}[tp]
\centering
\caption{Verified scores by capability and model (\%), with 95\% confidence intervals from the episode bootstrap.}
\label{tab:app_verified}
\small
\setlength{\tabcolsep}{3.2pt}
\renewcommand{\arraystretch}{0.99}
\psycheapptablefit{%
\begin{tabular}{lrrrrrrrr}
\toprule
Model & G & W & N & T & S & A & M & R \\
\midrule
Claude Opus 5 & \shortstack{89.31\\{\scriptsize [85.25,93.14]}} & \shortstack{77.80\\{\scriptsize [73.35,81.97]}} & \shortstack{69.10\\{\scriptsize [60.99,77.02]}} & \shortstack{40.70\\{\scriptsize [30.98,50.52]}} & \shortstack{79.18\\{\scriptsize [71.15,86.72]}} & \shortstack{86.24\\{\scriptsize [79.70,92.15]}} & \shortstack{86.56\\{\scriptsize [79.45,93.57]}} & \shortstack{84.82\\{\scriptsize [77.20,91.78]}} \\
Gemini 3.1 Pro & \shortstack{91.30\\{\scriptsize [87.47,94.82]}} & \shortstack{76.26\\{\scriptsize [72.01,80.50]}} & \shortstack{60.29\\{\scriptsize [51.65,68.84]}} & \shortstack{44.71\\{\scriptsize [35.77,53.49]}} & \shortstack{68.22\\{\scriptsize [59.22,76.84]}} & \shortstack{76.80\\{\scriptsize [69.16,83.91]}} & \shortstack{90.49\\{\scriptsize [84.00,96.14]}} & \shortstack{86.57\\{\scriptsize [78.99,93.24]}} \\
Gemini 3.6 Flash & \shortstack{91.76\\{\scriptsize [88.02,95.11]}} & \shortstack{79.68\\{\scriptsize [75.56,83.63]}} & \shortstack{69.97\\{\scriptsize [62.36,77.55]}} & \shortstack{47.81\\{\scriptsize [38.13,57.45]}} & \shortstack{79.32\\{\scriptsize [72.05,86.16]}} & \shortstack{89.58\\{\scriptsize [83.99,94.62]}} & \shortstack{90.92\\{\scriptsize [84.31,96.13]}} & \shortstack{91.21\\{\scriptsize [84.47,96.88]}} \\
GLM-5V-Turbo & \shortstack{89.18\\{\scriptsize [84.93,93.09]}} & \shortstack{76.72\\{\scriptsize [72.49,80.78]}} & \shortstack{61.82\\{\scriptsize [53.35,70.16]}} & \shortstack{34.23\\{\scriptsize [25.17,43.43]}} & \shortstack{68.14\\{\scriptsize [59.22,76.27]}} & \shortstack{79.87\\{\scriptsize [72.71,86.66]}} & \shortstack{76.23\\{\scriptsize [64.51,86.73]}} & \shortstack{83.37\\{\scriptsize [75.32,90.71]}} \\
GPT-5.6 Sol & \shortstack{92.13\\{\scriptsize [88.41,95.47]}} & \shortstack{78.75\\{\scriptsize [74.52,82.80]}} & \shortstack{67.28\\{\scriptsize [58.73,75.56]}} & \shortstack{32.99\\{\scriptsize [24.05,42.11]}} & \shortstack{74.86\\{\scriptsize [66.73,82.36]}} & \shortstack{84.25\\{\scriptsize [77.10,90.65]}} & \shortstack{91.11\\{\scriptsize [84.07,97.02]}} & \shortstack{86.47\\{\scriptsize [78.71,93.28]}} \\
Kimi K3 & \shortstack{91.52\\{\scriptsize [87.60,95.13]}} & \shortstack{76.11\\{\scriptsize [71.63,80.39]}} & \shortstack{60.62\\{\scriptsize [51.89,69.07]}} & \shortstack{38.82\\{\scriptsize [28.92,48.54]}} & \shortstack{79.38\\{\scriptsize [71.62,86.31]}} & \shortstack{84.28\\{\scriptsize [77.37,90.63]}} & \shortstack{85.17\\{\scriptsize [77.57,92.08]}} & \shortstack{86.87\\{\scriptsize [79.55,93.31]}} \\
Qwen3.8-Max & \shortstack{90.87\\{\scriptsize [86.80,94.49]}} & \shortstack{77.61\\{\scriptsize [73.21,81.77]}} & \shortstack{69.17\\{\scriptsize [61.13,76.74]}} & \shortstack{39.33\\{\scriptsize [29.58,49.19]}} & \shortstack{72.84\\{\scriptsize [64.08,80.86]}} & \shortstack{82.46\\{\scriptsize [75.42,88.98]}} & \shortstack{84.48\\{\scriptsize [76.62,91.83]}} & \shortstack{87.94\\{\scriptsize [80.62,94.32]}} \\
Seed 2.1 Pro & \shortstack{88.53\\{\scriptsize [84.27,92.60]}} & \shortstack{76.87\\{\scriptsize [72.73,80.88]}} & \shortstack{53.24\\{\scriptsize [44.28,62.13]}} & \shortstack{32.93\\{\scriptsize [23.95,42.46]}} & \shortstack{63.27\\{\scriptsize [54.27,71.82]}} & \shortstack{78.09\\{\scriptsize [70.74,85.13]}} & \shortstack{82.42\\{\scriptsize [73.06,90.96]}} & \shortstack{81.98\\{\scriptsize [73.57,89.66]}} \\
Qwen3.8-27B & \shortstack{90.93\\{\scriptsize [87.19,94.40]}} & \shortstack{80.53\\{\scriptsize [76.46,84.43]}} & \shortstack{66.96\\{\scriptsize [58.66,75.05]}} & \shortstack{28.58\\{\scriptsize [20.34,37.13]}} & \shortstack{68.87\\{\scriptsize [60.78,76.59]}} & \shortstack{83.01\\{\scriptsize [75.95,89.52]}} & \shortstack{77.77\\{\scriptsize [67.09,87.73]}} & \shortstack{85.27\\{\scriptsize [78.14,91.67]}} \\
Qwen3.5-9B & \shortstack{90.24\\{\scriptsize [86.41,93.86]}} & \shortstack{76.86\\{\scriptsize [72.51,81.10]}} & \shortstack{54.66\\{\scriptsize [46.32,62.78]}} & \shortstack{29.13\\{\scriptsize [21.11,37.16]}} & \shortstack{62.63\\{\scriptsize [53.85,70.87]}} & \shortstack{73.45\\{\scriptsize [65.10,81.29]}} & \shortstack{68.14\\{\scriptsize [56.45,78.92]}} & \shortstack{76.52\\{\scriptsize [68.04,84.25]}} \\
InternVL3.5-14B & \shortstack{63.36\\{\scriptsize [56.67,69.91]}} & \shortstack{56.63\\{\scriptsize [51.62,61.59]}} & \shortstack{19.90\\{\scriptsize [13.59,26.60]}} & \shortstack{15.63\\{\scriptsize [10.40,21.18]}} & \shortstack{34.05\\{\scriptsize [24.70,43.50]}} & \shortstack{45.10\\{\scriptsize [36.22,54.49]}} & \shortstack{59.53\\{\scriptsize [46.31,71.92]}} & \shortstack{63.22\\{\scriptsize [52.53,73.68]}} \\
Gemma-4-12B-it & \shortstack{79.48\\{\scriptsize [74.12,84.62]}} & \shortstack{71.19\\{\scriptsize [66.56,75.65]}} & \shortstack{46.41\\{\scriptsize [37.97,54.88]}} & \shortstack{13.49\\{\scriptsize [7.82,19.57]}} & \shortstack{49.43\\{\scriptsize [40.52,58.19]}} & \shortstack{62.32\\{\scriptsize [53.53,70.86]}} & \shortstack{65.07\\{\scriptsize [52.75,77.17]}} & \shortstack{74.58\\{\scriptsize [65.61,83.09]}} \\
InternVL3.5-8B & \shortstack{72.30\\{\scriptsize [66.50,77.95]}} & \shortstack{64.39\\{\scriptsize [59.56,69.13]}} & \shortstack{35.38\\{\scriptsize [28.74,42.35]}} & \shortstack{20.65\\{\scriptsize [13.52,28.42]}} & \shortstack{50.37\\{\scriptsize [41.59,58.77]}} & \shortstack{59.32\\{\scriptsize [50.99,67.60]}} & \shortstack{72.20\\{\scriptsize [61.99,82.48]}} & \shortstack{68.67\\{\scriptsize [60.04,77.10]}} \\
Phi-4-multimodal & \shortstack{52.50\\{\scriptsize [46.11,58.91]}} & \shortstack{50.32\\{\scriptsize [45.40,55.34]}} & \shortstack{22.14\\{\scriptsize [15.50,29.05]}} & \shortstack{20.85\\{\scriptsize [14.47,27.54]}} & \shortstack{45.96\\{\scriptsize [37.42,54.32]}} & \shortstack{58.17\\{\scriptsize [49.12,67.09]}} & \shortstack{64.63\\{\scriptsize [52.47,76.26]}} & \shortstack{74.44\\{\scriptsize [64.69,83.70]}} \\
InternVL3.5-4B & \shortstack{72.46\\{\scriptsize [66.80,77.81]}} & \shortstack{60.06\\{\scriptsize [55.26,64.74]}} & \shortstack{26.90\\{\scriptsize [20.36,33.68]}} & \shortstack{21.61\\{\scriptsize [14.07,29.38]}} & \shortstack{37.44\\{\scriptsize [28.78,46.41]}} & \shortstack{47.00\\{\scriptsize [38.29,55.57]}} & \shortstack{63.74\\{\scriptsize [51.08,76.30]}} & \shortstack{62.54\\{\scriptsize [52.94,71.94]}} \\
MiniCPM-V4.5 & \shortstack{81.26\\{\scriptsize [76.13,86.03]}} & \shortstack{70.63\\{\scriptsize [66.31,74.79]}} & \shortstack{33.81\\{\scriptsize [26.69,41.14]}} & \shortstack{24.32\\{\scriptsize [16.99,32.11]}} & \shortstack{50.08\\{\scriptsize [41.17,59.08]}} & \shortstack{62.45\\{\scriptsize [54.25,70.42]}} & \shortstack{65.40\\{\scriptsize [53.53,76.81]}} & \shortstack{77.13\\{\scriptsize [68.28,85.37]}} \\
Qwen3.5-4B & \shortstack{60.70\\{\scriptsize [53.98,67.23]}} & \shortstack{46.49\\{\scriptsize [41.21,51.94]}} & \shortstack{34.18\\{\scriptsize [26.60,41.90]}} & \shortstack{23.17\\{\scriptsize [15.95,30.83]}} & \shortstack{24.02\\{\scriptsize [16.73,31.61]}} & \shortstack{35.85\\{\scriptsize [27.88,44.11]}} & \shortstack{33.61\\{\scriptsize [20.47,47.54]}} & \shortstack{43.66\\{\scriptsize [33.09,54.56]}} \\
\bottomrule
\end{tabular}}
\end{table*}

\subsection{Confidence Intervals}
\label{app:serving_failures}
Table~\ref{tab:app_main_uncertainty} reports confidence intervals
for the overall Raw and Verified scores.
All scores use the full 1,000-question denominator.
\begin{table}[H]
\centering
\caption{Overall 95\% confidence intervals from bootstrapping episodes (\%). All models are evaluated on 1,000 questions.}
\label{tab:app_main_uncertainty}
\small
\setlength{\tabcolsep}{4pt}
\renewcommand{\arraystretch}{1.08}
\psycheapptablefit{%
\begin{tabular}{lrr}
\toprule
Model & Raw CI & Verified CI \\
\midrule
Claude Opus 5 & [79.75, 84.83] & [74.15, 79.22] \\
Gemini 3.1 Pro & [78.68, 83.96] & [71.75, 76.89] \\
Gemini 3.6 Flash & [83.42, 87.84] & [77.71, 82.27] \\
GLM-5V-Turbo & [75.58, 81.11] & [68.42, 73.94] \\
GPT-5.6 Sol & [78.53, 83.59] & [73.46, 78.45] \\
Kimi K3 & [77.94, 83.16] & [72.76, 77.85] \\
Qwen3.8-Max & [78.95, 83.99] & [73.04, 78.11] \\
Seed 2.1 Pro & [73.53, 79.17] & [66.97, 72.35] \\
Qwen3.8-27B & [77.26, 82.60] & [70.10, 75.33] \\
Qwen3.5-9B & [73.53, 79.43] & [63.58, 69.22] \\
InternVL3.5-14B & [48.83, 55.45] & [41.60, 47.67] \\
Gemma-4-12B-it & [63.30, 69.42] & [54.81, 60.65] \\
InternVL3.5-8B & [64.45, 70.29] & [52.65, 58.19] \\
Phi-4-multimodal & [57.84, 64.21] & [45.76, 51.52] \\
InternVL3.5-4B & [58.13, 64.46] & [46.00, 51.94] \\
MiniCPM-V4.5 & [66.55, 72.67] & [55.29, 60.97] \\
Qwen3.5-4B & [43.96, 51.42] & [34.52, 40.92] \\
\bottomrule
\end{tabular}}
\end{table}

\section{Conditional Rationale and Stratified Analysis}
\label{app:foundation_audit}

\subsection{Timing: Answer Credit and Evidence Support}
\label{app:timing_verification}

Timing has the lowest Verified score and credit retention for all
17 frozen models. Its 72 questions are all single-choice and come
from 72 episodes. Restricting every capability to single-choice
questions preserves the pattern of lowest retention for 17/17 models;
the pattern of lowest Verified scores holds for 16/17. Thus, Timing's difficulty
is reflected in both the absolute score and the credit retained after
rationale verification.

Table~\ref{tab:app_conditional_support} conditions on full answer
credit and holds answer format fixed. Across the 17 models, 569
correct Timing responses retain 64.40\% of answer credit
(95\% CI from bootstrapping episodes $[58.64,69.59]$).
Weak/contradicted Evidence Support labels occur in 283/569 responses (49.74\%),
compared with 837/10,988 (7.62\%) for other correct single-choice
responses: a difference of 42.12 points ($[33.37,51.50]$).
Weak/contradictory logic occurs in 182/569 (31.99\%) versus
561/10,988 (5.11\%), a difference of 26.88 points ($[19.29,35.27]$).
\begin{table}[!htbp]
\centering
\caption{Conditional verification across all 17 frozen models, restricted to correct single-choice answers. $n$ counts individual model responses to questions. Retention is mean Verified credit given Raw $=1$; adverse Evidence Support/Logical Integrity labels include weak and contradicted/contradictory. Rates are percentages.}
\label{tab:app_conditional_support}
\small
\setlength{\tabcolsep}{4pt}
\renewcommand{\arraystretch}{1.08}
\psycheapptablefit{%
\begin{tabular}{lrrrr}
\toprule
Cap. & $n$ & Retention & \shortstack{Weak\\support} & \shortstack{Weak\\logic} \\
\midrule
G & 2686 & 94.97 & 5.96 & 3.91 \\
W & 3728 & 95.30 & 4.24 & 3.00 \\
N & 672 & 84.78 & 18.30 & 14.58 \\
T & 569 & 64.40 & 49.74 & 31.99 \\
S & 800 & 92.72 & 6.50 & 5.12 \\
A & 1197 & 89.59 & 11.19 & 7.35 \\
M & 866 & 88.57 & 12.47 & 8.78 \\
R & 1039 & 92.41 & 9.82 & 3.95 \\
\bottomrule
\end{tabular}}
\end{table}

All 182 Timing responses with weak/contradictory logic also have
weak/contradicted Evidence Support labels; 101 have only the adverse Evidence Support label,
and 286 have neither adverse label. This overlap indicates that
the two judgments often identify problems in the same explanation.
Within the eight proprietary models, the difference in Evidence Support labels remains
35.60 points ($[25.79,46.23]$): 123/316 correct Timing responses
versus 187/5,633 other correct single-choice responses.

\subsection{Need Answer Completeness}
\label{app:need_profiles}
Claude and Gemini have similar aggregate Need Verified scores,
69.10 and 69.97, but their ordering differs between choice and open
questions (Table~\ref{tab:app_need_formats}). Claude trails Gemini by
10.36 points on 62 choice questions (95\% CI $[-20.94,-0.31]$)
and leads by 15.95 points on 35 open questions ($[4.36,26.74]$).
These subsets differ in both question content and answer requirements.
\begin{table}[!htbp]
\centering
\caption{Need scores by question format (\%), followed by credit counts on the same 35 open questions.}
\label{tab:app_need_formats}
\small
\setlength{\tabcolsep}{4pt}
\renewcommand{\arraystretch}{1.08}
\psycheapptablefit{%
\begin{tabular}{lrrrr}
\toprule
Model & \multicolumn{2}{c}{Choice} & \multicolumn{2}{c}{Open} \\
\cmidrule(lr){2-3}\cmidrule(lr){4-5}
 & Raw & Ver. & Raw & Ver. \\
\midrule
Claude Opus 5 & 75.81 & 70.95 & 78.57 & 65.83 \\
Gemini 3.1 Pro & 79.03 & 72.51 & 47.14 & 38.65 \\
Gemini 3.6 Flash & 87.10 & 81.31 & 57.14 & 49.88 \\
GLM-5V-Turbo & 74.19 & 63.94 & 65.71 & 58.06 \\
GPT-5.6 Sol & 74.19 & 71.38 & 68.57 & 60.00 \\
Kimi K3 & 66.13 & 62.15 & 71.43 & 57.90 \\
Qwen3.8-Max & 79.03 & 73.53 & 71.43 & 61.43 \\
Seed 2.1 Pro & 59.68 & 54.77 & 58.57 & 50.52 \\
\bottomrule
\end{tabular}}
\par\smallskip
\psycheapptablefit{%
\begin{tabular}{lrrr}
\toprule
Open Need & Raw $=0$ & Raw $=0.5$ & Raw $=1$ \\
\midrule
Gemini 3.6 Flash & 5 & 20 & 10 \\
Claude Opus 5 & 4 & 7 & 24 \\
\bottomrule
\end{tabular}}
\end{table}

On the same 35 open Need questions, Gemini receives 5/20/10
judgments assigning zero, partial, or full credit; Claude receives 4/7/24.
Their rates of nonzero credit are 85.71/88.57\%, while rates of full credit are 28.57/68.57\%. The rubric permits an exact decomposition:
\begin{equation}
\mathbb E[a]=0.5P(a>0)+0.5P(a=1).
\end{equation}
Claude's Raw advantage of 21.43 points comprises 1.43 points from
coverage with nonzero credit and 20.00 from coverage with full credit.
The difference in full credit is 40.00 points ($[18.42,60.87]$).
Seventeen questions move from Gemini partial to Claude full credit,
compared with three in the reverse direction. The decomposition
identifies answer completeness as the main source of the Raw gap.

\subsection{Answer Completeness and Rationale Support}
\label{app:answer_completeness}
Table~\ref{tab:app_open_diagnostics} separates credit for open answers
from length of the final answer. Claude averages 52.3 words, compared
with Gemini's 16.2; their open Raw scores are 76.36 and 67.27.

Among the 110 open questions, 38 receive full answer credit from
both Claude and Gemini. On these jointly correct questions, their mean Verified credits are
90.42 and 97.15, respectively. Comparing jointly correct answers
distinguishes completeness of the final answer from rationale quality.
\begin{table}[!htbp]
\centering
\caption{Diagnostics for open answers on 110 questions. Raw and Verified are micro means (\%), including partial answer credit. Words denotes the word count of the final answer field.}
\label{tab:app_open_diagnostics}
\small
\setlength{\tabcolsep}{4pt}
\renewcommand{\arraystretch}{1.08}
\psycheapptablefit{%
\begin{tabular}{lrrrrr}
\toprule
Model & Raw & Ver. & Words & Full & Partial \\
\midrule
Claude Opus 5 & 76.36 & 64.83 & 52.3 & 72 & 24 \\
Gemini 3.1 Pro & 49.09 & 42.31 & 14.7 & 28 & 52 \\
Gemini 3.6 Flash & 67.27 & 60.18 & 16.2 & 51 & 46 \\
GLM-5V-Turbo & 63.18 & 54.14 & 18.5 & 49 & 41 \\
GPT-5.6 Sol & 70.00 & 62.71 & 19.6 & 59 & 36 \\
Kimi K3 & 69.09 & 59.02 & 26.2 & 59 & 34 \\
Qwen3.8-Max & 66.36 & 57.30 & 21.7 & 56 & 34 \\
Seed 2.1 Pro & 57.73 & 47.82 & 15.8 & 41 & 45 \\
\bottomrule
\end{tabular}}
\end{table}

\subsection{Paired Rationale and Weight Sensitivity}
\label{app:weight_sensitivity}

Qwen3.8-27B and Qwen3.5-9B jointly receive full answer credit
on 721 questions from 480 episodes.
Their mean verification weights are 95.11 and 92.26,
with a paired difference of 2.84 points ($[1.66,4.05]$).
The paired difference measures rationale quality on jointly
correct answers.
Alternative Soft and Strict weights retain the ordering
Qwen3.5-9B $>$ MiniCPM $>$ InternVL
(Table~\ref{tab:app_scoring_sensitivity});
the underlying Judge labels remain fixed.
For all 17 frozen models, Timing remains the capability with the lowest retention under the current product, mean or minimum of normalized
quality factors, weights using only support or only logic, and a binary
threshold requiring at least normal support and acceptable logic.
Requiring both strongest labels preserves the pattern for 16/17.
\begin{table}[!htbp]
\centering
\caption{Verified macro scores under alternative weights (\%). Soft uses 1.1/1.0/0.9/0.7; Strict uses 1.2/1.0/0.6/0.3; each is normalized by its maximum product. The second panel counts frozen models whose Timing retention is lowest under alternative aggregations.}
\label{tab:app_scoring_sensitivity}
\small
\setlength{\tabcolsep}{4pt}
\renewcommand{\arraystretch}{1.08}
\psycheapptablefit{%
\begin{tabular}{lrrr}
\toprule
Model & Soft & Main & Strict \\
\midrule
Qwen3.5-9B & 70.36 & 66.45 & 64.73 \\
MiniCPM-V4.5 & 62.63 & 58.13 & 55.91 \\
InternVL3.5-8B & 60.17 & 55.41 & 53.32 \\
\bottomrule
\end{tabular}}
\par\smallskip
\psycheapptablefit{%
\begin{tabular}{lr}
\toprule
Aggregation & Timing lowest \\
\midrule
Current product & 17/17 \\
Mean of normalized factors & 17/17 \\
Minimum normalized factor & 17/17 \\
Support only & 17/17 \\
Logic only & 17/17 \\
Both at least normal/acceptable & 17/17 \\
Both strongest & 16/17 \\
\bottomrule
\end{tabular}}
\end{table}

\subsection{Question Format, Source, and Duration}
\label{app:stratified_analysis}
Table~\ref{tab:app_formats} reports micro scores for 890 choice
questions and 110 open questions. The choice group combines all
questions answered by selecting options under the same exact matching
rule (Appendix~\ref{app:scoring_details}).
\begin{table*}[!t]
\centering
\caption{Micro scores by question format (\%). Choice and open groups contain 890 and 110 questions, respectively.}
\label{tab:app_formats}
\small
\setlength{\tabcolsep}{4pt}
\renewcommand{\arraystretch}{1.08}
\psycheapptablefit{%
\begin{tabular}{lrrrr}
\toprule
Model & Choice Raw & Choice Ver. & Open Raw & Open Ver. \\
\midrule
Claude Opus 5 & 83.48 & 80.08 & 76.36 & 64.83 \\
Gemini 3.1 Pro & 85.51 & 80.31 & 49.09 & 42.31 \\
Gemini 3.6 Flash & 87.75 & 83.70 & 67.27 & 60.18 \\
GLM-5V-Turbo & 82.02 & 76.96 & 63.18 & 54.14 \\
GPT-5.6 Sol & 83.60 & 80.21 & 70.00 & 62.71 \\
Kimi K3 & 82.47 & 79.46 & 69.09 & 59.02 \\
Qwen3.8-Max & 83.82 & 80.19 & 66.36 & 57.30 \\
Seed 2.1 Pro & 81.24 & 76.21 & 57.73 & 47.82 \\
Qwen3.5-9B & 81.80 & 74.83 & 56.82 & 44.58 \\
InternVL3.5-8B & 72.13 & 62.31 & 43.18 & 30.83 \\
MiniCPM-V4.5 & 76.85 & 67.12 & 41.36 & 30.49 \\
\bottomrule
\end{tabular}}
\end{table*}

We also group questions by source and video duration
(Tables~\ref{tab:app_sources} and~\ref{tab:app_strata}).
Qwen's Raw micro score is 87.69 on YouTube Vlog and 67.52
on YouTube Interview. Sources differ in task composition;
IndEgo contains only eight questions.
Duration groups contain 555/303/142 questions at
$\leq30$/30-60/$>60$ seconds. Table~\ref{tab:app_strata} also
reports the 110 open questions.
\begin{table*}[!t]
\centering
\caption{Micro scores by source (Raw/Verified, \%) for the three diagnostic models.}
\label{tab:app_sources}
\small
\setlength{\tabcolsep}{4pt}
\renewcommand{\arraystretch}{1.08}
\psycheapptablefit{%
\begin{tabular}{lrrrr}
\toprule
Source & n & Qwen & InternVL & MiniCPM \\
\midrule
CASTLE & 87 & 73.56/69.56 & 63.22/57.96 & 68.39/62.35 \\
EasyCom & 70 & 69.29/57.07 & 65.71/57.02 & 63.57/53.54 \\
Ego4D & 230 & 75.22/65.35 & 62.17/51.50 & 63.70/52.80 \\
EgoCom & 60 & 85.00/78.47 & 70.83/61.11 & 77.50/65.71 \\
IndEgo & 8 & 62.50/41.06 & 62.50/53.12 & 75.00/54.51 \\
MERCI+ & 91 & 81.87/73.07 & 74.73/61.92 & 70.88/63.28 \\
YouTube Interview & 117 & 67.52/64.13 & 63.68/53.87 & 68.80/59.94 \\
YouTube Vlog & 337 & 87.69/80.82 & 75.82/65.11 & 83.53/73.08 \\
\bottomrule
\end{tabular}}
\end{table*}

\begin{table*}[!t]
\centering
\caption{Strata defined by metadata: micro Raw/Verified scores (\%). Open denotes the open question subset.}
\label{tab:app_strata}
\small
\setlength{\tabcolsep}{4pt}
\renewcommand{\arraystretch}{1.08}
\psycheapptablefit{%
\begin{tabular}{lrrrr}
\toprule
Subset & n & Qwen & InternVL & MiniCPM \\
\midrule
Open & 110 & 56.82/44.58 & 43.18/30.83 & 41.36/30.49 \\
$\leq 30$ s & 555 & 76.49/69.78 & 65.59/56.24 & 70.45/62.13 \\
30-60 s & 303 & 82.51/74.45 & 72.61/61.17 & 76.07/65.11 \\
$>60$ s & 142 & 81.69/71.93 & 74.30/64.10 & 76.06/62.57 \\
\bottomrule
\end{tabular}}
\end{table*}

\section{Complementary Answers and Voting}
\label{app:failure_analysis}

Among 72 T single-choice questions, 15 are answered correctly
by all eight proprietary models and six are answered incorrectly by all.
The best Raw score from a single model is 48/72 (66.67\%), whereas
selecting an existing correct answer with the reference yields
66/72 (91.67\%; Table~\ref{tab:app_closed_shared}).
The oracle using reference answers bounds recoverable answer coverage.
For each question, each selected option ID casts one vote. Plurality selects the
option with the most votes. When $r$ options tie, expected answer credit is
$1/r$ if the correct option is among them and zero otherwise.
Across all 72 Timing questions, including three with tied highest
votes, expected plurality accuracy is 41.5/72 (57.64\%) under uniform
tie breaking.

The Oracle gain over Gemini is $+25.00$ points (95\% CI $[15.58,35.48]$):
18 of Gemini's 24 incorrect answers have a correct answer from another model.
The Plurality difference from Gemini is $-9.03$ points ($[-20.00,1.54]$),
with the interval spanning zero. At least five of eight models choose the same wrong
option on 21 questions, 18 of which still have a correct candidate.
Strict majority voting (at least five agreeing models, otherwise
abstention with zero credit) reaches 36/72 (50.00\%).
These contrasts distinguish candidate coverage from successful
selection.

On the 69 T questions without a tie for the most votes, plurality
accuracy is 59.42\%, below Gemini's 68.12\% on the same questions.
Model disagreement contains recoverable information, but simple
voting does not reliably identify the correct response.
\begin{table}[!htbp]
\centering
\caption{Answer overlap and plurality voting for eight proprietary models on single-choice questions. Best, Oracle, and Vote are Raw percentages. Vote uses uniform tie breaking; Oracle identifies any correct existing answer with the reference and is an upper bound.}
\label{tab:app_closed_shared}
\small
\setlength{\tabcolsep}{4pt}
\renewcommand{\arraystretch}{1.08}
\psycheapptablefit{%
\begin{tabular}{lrrrrrr}
\toprule
Cap. & $n$ & \shortstack{All\\correct} & \shortstack{All\\wrong} & Best & Oracle & Vote \\
\midrule
all & 858 & 595 & 20 & 90.91 & 97.67 & 89.97 \\
G & 181 & 145 & 1 & 95.58 & 99.45 & 96.59 \\
W & 258 & 200 & 4 & 92.25 & 98.45 & 92.96 \\
N & 62 & 24 & 2 & 87.10 & 96.77 & 82.66 \\
T & 72 & 15 & 6 & 66.67 & 91.67 & 57.64 \\
S & 64 & 43 & 3 & 90.62 & 95.31 & 85.94 \\
A & 87 & 60 & 2 & 95.40 & 97.70 & 94.83 \\
M & 61 & 50 & 0 & 98.36 & 100.00 & 96.72 \\
R & 73 & 58 & 2 & 93.15 & 97.26 & 93.15 \\
\bottomrule
\end{tabular}}
\end{table}

\section{Multimodal Cues and Input Controls}
\label{app:modality_analysis}
\label{app:ablation}

\subsection{Modality Ablations}
\label{app:modality_ablation_details}

\begin{figure*}[tp]
\centering
\includegraphics[width=\textwidth]{\PsycheExpRoot/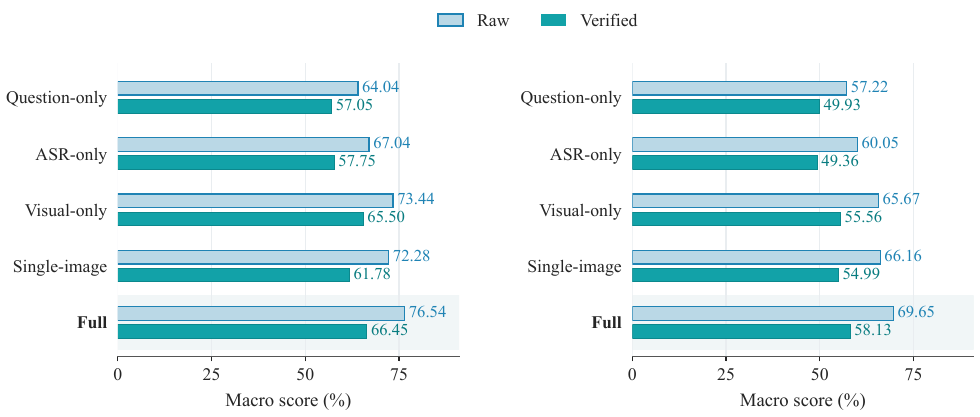}
\caption{Input controls for Qwen3.5-9B (left) and MiniCPM-V4.5 (right) on the same 1,000 PsycheMME questions. Paired bars show Raw and Verified macro scores. Full supplies video and ASR; Question-only supplies neither; ASR-only omits frames; Visual-only omits ASR. Single-image replaces the video frames with one midpoint image while retaining ASR. The Full condition is shaded for reference.}
\label{fig:app_modality_comparison}
\end{figure*}

We compare four input conditions on the same 1,000 questions:
Full, Question-only, ASR-only, and Visual-only. Question-only uses
the question and answer options; ASR-only supplies transcripts
without video frames; Visual-only supplies frames without ASR.
Table~\ref{tab:app_modality} reports macro scores and differences
relative to Full. Figure~\ref{fig:app_modality_comparison} summarizes
these conditions together with the single image control below.

Full exceeds Question-only by 12.50 Raw points for Qwen and
12.43 for MiniCPM. Visual-only reaches 73.44 and 65.67, respectively,
while Full reaches 76.54 and 69.65. The corresponding increases
over Visual-only are 3.10 and 3.98 points.
\begin{table}[!htbp]
\centering
\caption{Modality macro scores (\%) and Raw changes relative to each model's Full input (pp).}
\label{tab:app_modality}
\small
\setlength{\tabcolsep}{4pt}
\renewcommand{\arraystretch}{1.08}
\psycheapptablefit{%
\begin{tabular}{llrrr}
\toprule
Model & Input & Raw & Ver. & $\Delta$ Raw \\
\midrule
Qwen3.5-9B & Full & 76.54 & 66.45 & - \\
 & Question-only & 64.04 & 57.05 & -12.50 \\
 & ASR-only & 67.04 & 57.75 & -9.50 \\
 & Visual-only & 73.44 & 65.50 & -3.10 \\
MiniCPM-V4.5 & Full & 69.65 & 58.13 & - \\
 & Question-only & 57.22 & 49.93 & -12.43 \\
 & ASR-only & 60.05 & 49.36 & -9.60 \\
 & Visual-only & 65.67 & 55.56 & -3.98 \\
\bottomrule
\end{tabular}}
\end{table}

\paragraph{Input benefits for each capability.}
Full exceeds ASR-only in G Raw by 30.65 points for Qwen and
45.16 for MiniCPM. Full exceeds Visual-only in Qwen N by 10.31
and MiniCPM S by 13.40 points.

\subsection{Single Images}
\label{app:frame_sampling}
\label{app:input_ablation}

The comparison using a single image replaces the sampled video frames with
one midpoint image while keeping the remaining inputs fixed.
Qwen Raw decreases by 4.26 macro points
(95\% CI $[-6.31,-2.26]$), including 12.63 points in G and 6.41 in W.
Its T score decreases by 8.33 points ($[-18.57,1.56]$).
MiniCPM Raw decreases by 3.49 macro points ($[-5.65,-1.31]$).
The aggregate losses indicate a benefit from broader visual coverage.

\section{Historical Context and Memory Controls}
\label{app:memory_analysis}

The history intervention evaluates Qwen on 40 Memory single-choice
questions from 24 episodes. Matched History supplies the relevant
history for each question; No History omits it; Mismatched History
substitutes another history using seeds 42, 2026, and 3407.
The video, question, and remaining inputs are held fixed.

\begin{figure}[!htbp]
\centering
\includegraphics[width=\columnwidth]{\PsycheExpRoot/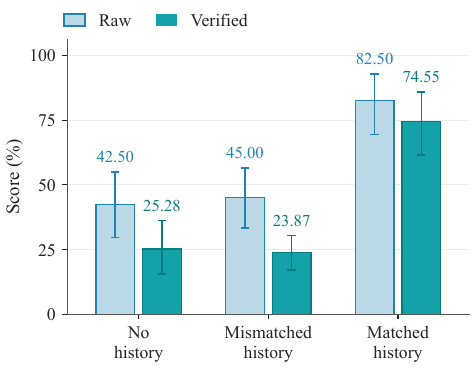}
\caption{History controls for Qwen3.5-9B on the same 40 Memory questions from 24 episodes. Bars show Raw and Verified scores with 95\% intervals from bootstrapping episodes. Mismatched History averages seeds 42, 2026, and 3407; its interval is computed for that mean.}
\label{fig:app_memory_context}
\end{figure}

Matched History achieves 82.50 Raw and 74.55 Verified, compared with
42.50/25.28 for No History. The paired gains are 40.00 Raw points
(95\% CI $[26.32,53.66]$) and 49.27 Verified points
($[35.80,61.97]$). Mismatched History averages 45.00/23.87 across
the three seeds (Table~\ref{tab:app_memory}).
Figure~\ref{fig:app_memory_context} compares these conditions and their
confidence intervals.

\begin{table}[H]
\centering
\caption{History interventions for Qwen on 40 Memory single-choice questions from 24 episodes. Raw and Verified are micro scores (\%); mismatched histories are sampled using three random seeds.}
\label{tab:app_memory}
\small
\setlength{\tabcolsep}{4pt}
\renewcommand{\arraystretch}{1.08}
\psycheapptablefit{%
\begin{tabular}{lrrrr}
\toprule
Condition & Seed & n & Raw & Ver. \\
\midrule
Matched history & - & 40 & 82.50 & 74.55 \\
No history & - & 40 & 42.50 & 25.28 \\
Mismatched history & 42 & 40 & 45.00 & 24.64 \\
Mismatched history & 2026 & 40 & 42.50 & 22.05 \\
Mismatched history & 3407 & 40 & 47.50 & 24.91 \\
Mean with mismatched history & 3 seeds & 40 & 45.00 & 23.87 \\
\bottomrule
\end{tabular}}
\end{table}

Relative to No History, Mismatched History changes Raw by
$+2.50$ points ($[-2.86,9.40]$) and Verified by $-1.41$
($[-9.13,5.66]$).

\section{Human Baseline}
\label{app:human_evaluation}

\subsection{Question Answering Protocol}
\label{app:human_baseline_protocol}

Five annotators independently answer 300 randomly sampled benchmark questions spanning all eight capabilities,
providing one response per annotator for each question (1,500 responses).
Answers are scored against the benchmark references. This experiment measures human answer accuracy.

\subsection{Aggregation and Reporting}
\label{app:human_baseline_reporting}

Raw micro accuracy pools all 1,500 responses: 1,279 are correct,
giving 85.27\%. Raw macro accuracy averages the pooled accuracy
within each of the eight capabilities, giving 84.82\%.
Table~\ref{tab:app_human_results} reports individual and pooled results;
Table~\ref{tab:app_human_capabilities} reports the capability breakdown.
\begin{table}[H]
\centering
\caption{Human answer accuracy. Five annotators independently answer 300 randomly sampled questions; micro pools responses, and macro averages the eight capability accuracies.}
\label{tab:app_human_results}
\small
\setlength{\tabcolsep}{5pt}
\renewcommand{\arraystretch}{1.10}
\psycheapptablefit{%
\begin{tabular}{lrrr}
\toprule
Annotator & Responses & Correct & Raw (\%) \\
\midrule
A & 300 & 260 & 86.67 \\
B & 300 & 253 & 84.33 \\
C & 300 & 259 & 86.33 \\
D & 300 & 254 & 84.67 \\
E & 300 & 253 & 84.33 \\
\midrule
Pooled micro & 1,500 & 1,279 & 85.27 \\
Capability macro & - & - & 84.82 \\
\bottomrule
\end{tabular}}
\end{table}

\begin{table}[H]
\centering
\caption{Human accuracy by capability. $n$ counts distinct questions; every question receives five independent answers.}
\label{tab:app_human_capabilities}
\small
\setlength{\tabcolsep}{5pt}
\renewcommand{\arraystretch}{1.10}
\psycheapptablefit{%
\begin{tabular}{lrrr}
\toprule
Capability & $n$ & Correct / responses & Raw (\%) \\
\midrule
G & 47 & 216/235 & 91.91 \\
W & 36 & 149/180 & 82.78 \\
N & 31 & 125/155 & 80.65 \\
T & 35 & 139/175 & 79.43 \\
S & 43 & 187/215 & 86.98 \\
A & 39 & 166/195 & 85.13 \\
M & 32 & 132/160 & 82.50 \\
R & 37 & 165/185 & 89.19 \\
\midrule
Micro & 300 & 1,279/1,500 & 85.27 \\
Macro & - & - & 84.82 \\
\bottomrule
\end{tabular}}
\end{table}

\section{Human-Judge Agreement}
\label{app:evaluator_robustness}
\label{app:human_endorsement}

We measure Human-Judge Agreement through expert endorsement of the
presented Judge decisions. Five psychology experts independently
review 300 randomly sampled Judge cases spanning all eight capabilities, yielding 1,500
assessments.
Each expert examines the Judge's labels and
justifications against the reference cues and records whether they
agree with the presented decision.

Experts endorse the Judge's decisions in 1,352 assessments (90.13\%).
The mean agreement rate across the eight capabilities is 89.79\%.
Table~\ref{tab:app_human_judge_final} reports individual and pooled
agreement, and Table~\ref{tab:app_human_judge_capabilities} reports
results by capability.
\begin{table}[H]
\centering
\caption{Human-Judge Agreement in 1,500 expert reviews of 300 randomly sampled Judge cases. Agreement denotes endorsement of the presented decision.}
\label{tab:app_human_judge_final}
\small
\setlength{\tabcolsep}{5pt}
\renewcommand{\arraystretch}{1.10}
\psycheapptablefit{%
\begin{tabular}{lrrr}
\toprule
Expert & Reviews & Endorsed & Rate (\%) \\
\midrule
A & 300 & 275 & 91.67 \\
B & 300 & 266 & 88.67 \\
C & 300 & 272 & 90.67 \\
D & 300 & 268 & 89.33 \\
E & 300 & 271 & 90.33 \\
\midrule
Pooled micro & 1,500 & 1,352 & 90.13 \\
Capability macro & - & - & 89.79 \\
\bottomrule
\end{tabular}}
\end{table}

\begin{table}[H]
\centering
\caption{Human-Judge Agreement by capability. $n$ counts distinct cases; every case receives five independent expert reviews.}
\label{tab:app_human_judge_capabilities}
\small
\setlength{\tabcolsep}{5pt}
\renewcommand{\arraystretch}{1.10}
\psycheapptablefit{%
\begin{tabular}{lrrr}
\toprule
Capability & $n$ & Endorsed / reviews & Rate (\%) \\
\midrule
G & 48 & 230/240 & 95.83 \\
W & 38 & 170/190 & 89.47 \\
N & 32 & 144/160 & 90.00 \\
T & 33 & 146/165 & 88.48 \\
S & 42 & 196/210 & 93.33 \\
A & 36 & 164/180 & 91.11 \\
M & 36 & 158/180 & 87.78 \\
R & 35 & 144/175 & 82.29 \\
\midrule
Micro & 300 & 1,352/1,500 & 90.13 \\
Macro & - & - & 89.79 \\
\bottomrule
\end{tabular}}
\end{table}

\section{Analysis Scope and Reproducibility}
\label{app:reproducibility}

\subsection{Sample Sizes and Evaluation Units}
\label{app:counting_units}

Scores on the full benchmark use all 1,000 questions. Additional analyses use the question,
episode, response, or review as their unit of observation.
Table~\ref{tab:app_analysis_scopes} summarizes these sample sizes.
Uncertainty estimation based on episodes retains the dependence among
questions from the same video and pairs the compared conditions.
\begin{table}[H]
\centering
\caption{Counting units and analysis scope.}
\label{tab:app_analysis_scopes}
\small
\setlength{\tabcolsep}{4pt}
\renewcommand{\arraystretch}{1.08}
\psycheapptablefit{%
\begin{tabular}{lrr}
\toprule
Analysis & Unit & Size \\
\midrule
Full evaluation & Question / episode & 1000 / 630 \\
Conditional rationale & Individual model response & Depends on capability \\
Proprietary Timing overlap & Single-choice question & 72 \\
Memory intervention & Question / episode & 40 / 24 \\
Human baseline & Question / response & 300 / 1500 \\
Human-Judge agreement & Judge case / review & 300 / 1500 \\
\bottomrule
\end{tabular}}
\end{table}

\subsection{Materials for Reproduction}
\label{app:result_provenance}

Recomputation requires the benchmark questions, model predictions,
Judge assessments, and evaluation configurations. Scores for individual
questions follow the rubric in Appendix~\ref{app:scoring_details}.
Each comparison uses the specified question subset, baseline condition,
and episode bootstrap procedure.

\end{document}